\pdfoutput=1

\documentclass[10pt,journal,compsoc,twoside]{IEEEtran}

\usepackage[utf8]{inputenc}
\usepackage{xspace}
\usepackage{amsmath}
\usepackage{amsfonts}
\usepackage{amssymb}
\usepackage{tikz}
\usepackage{pgfplots}
\usepackage{standalone}
\usepackage{tabularx}
\usepackage{array}
\usepackage{booktabs}
\usepackage{cite}
\usepackage{colortbl}
\usepackage{microtype}
\ifCLASSOPTIONcompsoc
  \usepackage[caption=false,font=footnotesize,labelfont=sf,textfont=sf]{subfig}
\else
	\usepackage[justification=centering, caption=false]{subfig}
\fi

\usepackage{algorithm}
\usepackage[noend]{algpseudocode}
\usepackage{enumitem}
\usepackage{calc}
\usepackage{threeparttable}
\usepackage{url}
\usepackage[hidelinks]{hyperref}
\usepackage{breakurl}
\newcolumntype{C}{>{\centering\arraybackslash}X}

\usetikzlibrary{calc}
\usetikzlibrary{fit}
\usetikzlibrary{shapes}
\usetikzlibrary{positioning}
\usetikzlibrary{matrix}

\newlength\eltWidth
\newlength\eltWidthb
\newlength\eltHeight
\def\horSep{0.5em}
\def\vertSep{\horSep}

\def\labelSep{2*\horSep}

\tikzstyle{elt} = [
  inner sep=0pt,
  outer sep=0pt,
  text width=\eltWidth,
  font=\scriptsize,
  align=center
]

\tikzstyle{eltb} = [
inner sep=0pt,
outer sep=0pt,
text width=\eltWidthb,
font=\scriptsize,
align=center
]

\tikzstyle{register} = [
  rectangle split,
  minimum height=\eltHeight,
  rectangle split horizontal=true,
  rectangle split parts=5,
  rectangle split part align=base,
  draw
]

\tikzstyle{operator} = [
  chamfered rectangle,
  font=\footnotesize,
  text width=13em,
  text height=1.2ex,
  text depth=0.1ex,
  align=center,
  draw
]

\tikzstyle{label} = [
    font=\footnotesize,
    inner sep=0,
    outer sep=0
]

\newlength\cellHeight
\newlength\cellWidth
\tikzset{
  memory cell/.style={
    rectangle,
    draw,
    font=\footnotesize,
    text depth=0.25ex,
    text height=\cellHeight,
    text width=\cellWidth,
    align=center,
  },      
  memory/.style={
    matrix of math nodes,
    row sep=-\pgflinewidth,
    column sep=-\pgflinewidth,
    nodes={
        memory cell
    },
    execute at empty cell={\node[draw=none]{};}
  },
  memory multicell/.style={
    draw,
    inner ysep=\pgflinewidth/2,
    inner xsep=-\pgflinewidth/2,
    font=\footnotesize,
    align=center,
    text height=\cellHeight,
    text depth=0.1ex,
  }
}

\tikzstyle{memory label} = [
    font=\footnotesize
]

\newlength\cellWidths
\tikzset{
	memorys cell/.style={
		rectangle,
		draw,
		font=\footnotesize,
		text depth=0.25ex,
		text height=\cellHeight,
		text width=\cellWidths,
		align=center,
	},      
	memorys/.style={
		matrix of math nodes,
		row sep=-\pgflinewidth,
		column sep=-\pgflinewidth,
		nodes={
			memorys cell
		},
		execute at empty cell={\node[draw=none]{};}
	},
	memorys multicell/.style={
		draw,
		inner ysep=\pgflinewidth/2,
		inner xsep=-\pgflinewidth/2,
		font=\footnotesize,
		align=center,
		text height=\cellHeight,
		text depth=0.1ex,
	}
}

\tikzstyle{memorys label} = [
font=\footnotesize
]

\newlength\cellWidthb
\tikzset{
	memory16 cell/.style={
		rectangle,
		draw,
		font=\footnotesize,
		text depth=0.25ex,
		text height=\cellHeight,
		text width=\cellWidthb,
		align=center,
	},      
	memory16/.style={
		matrix of math nodes,
		row sep=-\pgflinewidth,
		column sep=-\pgflinewidth,
		nodes={
			memory16 cell
		},
		execute at empty cell={\node[draw=none]{};}
	},
	memory16 multicell/.style={
		draw,
		inner ysep=\pgflinewidth/2,
		inner xsep=-\pgflinewidth/2,
		font=\footnotesize,
		align=center,
		text height=\cellHeight,
		text depth=0.1ex,
	}
}

\tikzstyle{memory16 label} = [
font=\footnotesize
]

\pgfplotsset{compat=1.12}
\pgfplotsset{small fonts/.style={
  title style = {font=\small, yshift=-0.5em},
  label style = {font=\small},
  tick label style = {font=\small},
  legend style={font=\small}
}}
\pgfplotsset{recall plot/.style={
  small fonts,
  xtick={1,2,5,10,20,50,100,200,500,1000},
  xticklabels={1,2,5,10,20,50,100,200,500,1K},
  xlabel=$R$,
  ylabel=Recall@$R$,
  legend pos=south east,
  legend cell align=left,
}}
\pgfplotsset{time plot/.style={
  small fonts,
  yticklabel=\empty,
  legend pos=south east,
  legend cell align=left,
  xlabel={Total query time [ms]},
  every node near coord/.append style={font=\small},
  nodes near coords align={horizontal},
  every node near coord/.append sex Akka team
  tyle={xshift=0.05em},
  nodes near coords={
        \pgfmathprintnumber[fixed relative, precision=2]{\pgfplotspointmeta}
  }
}}
\pgfplotsset{time bar/.style={
  xbar, point meta=explicit,
  error bars/.cd,
  x dir=both,
  x explicit
}}

\usepgfplotslibrary{groupplots}

\newcommand*{\eg}{e.g.,\xspace}
\newcommand*{\ie}{i.e.,\xspace}

\newcommand*{\stimes}{{\times}}
\newcommand*{\pq}[2]{PQ\,${#1}\stimes{#2}$}

\newcommand*{\q}[1][]{%
  \ifthenelse{\isempty{#1}}%
  {\operatorname{q}}%
  {\operatorname{q_#1}}%
}

\newcommand*{\uv}[1][]{%
  \ifthenelse{\isempty{#1}}%
  {\operatorname{u}}%
  {\operatorname{u_#1}}%
}

\DeclareMathOperator*{\argmin}{arg\,min}

\newcolumntype{Y}{>{\centering\arraybackslash}X}

\def\ipq#1#2{
	$#1\stimes\{#2\}$
}

\def\rpq#1#2{
	$#1\stimes#2$
}

\newenvironment{rev}{}{}
\newcommand{\trev}[1]{#1}

\begin{document}

\title{Quicker ADC : Unlocking the Hidden Potential of Product Quantization with SIMD}

\author{Fabien~André,
	Anne-Marie~Kermarrec,
	and~Nicolas~Le~Scouarnec
	\IEEEcompsocitemizethanks{%
	\IEEEcompsocthanksitem Fabien André and Nicolas Le Scouarnec were with Technicolor, France. %
	\IEEEcompsocthanksitem Anne-Marie Kermarrec is with Inria, France,  EPFL, Switzerland, and Mediego, France. %
	}
	\thanks{IEEE Transactions on Pattern Analysis and Machine Intelligence\protect\\ 
		\href{https://doi.org/10.1109/TPAMI.2019.2952606}{DOI: 10.1109/TPAMI.2019.2952606}\protect\\
		© 2019 IEEE.  Personal use of this material is permitted.  Permission from IEEE must be obtained for all other uses, in any current or future media, including reprinting/republishing this material for advertising or promotional purposes, creating new collective works, for resale or redistribution to servers or lists, or reuse of any copyrighted component of this work in other works.
		}}

\markboth{IEEE Transactions on Pattern Analysis and Machine Intelligence}%
{André \MakeLowercase{\textit{et al.}}: Quicker ADC : Unlocking the Hidden Potential of Product Quantization with SIMD}

%
%
%


\IEEEtitleabstractindextext{
	\begin{abstract}
%

Efficient Nearest Neighbor (NN) search in high-dimensional spaces is a foundation of many multimedia retrieval systems. A common approach is to rely on Product Quantization, which allows the storage of large vector databases in memory and efficient distance computations. Yet, implementations of nearest neighbor search with Product Quantization have their performance limited by the many memory accesses they perform. Following this observation, André et al. proposed Quick ADC with up to $6\stimes$ faster implementations of \pq{m}{4} product quantizers (PQ) leveraging specific SIMD instructions.

Quicker ADC is a generalization of Quick ADC not limited to \pq{m}{4} codes and supporting AVX-512, the latest revision of SIMD instruction set. In doing so, Quicker ADC faces the challenge of using efficiently 5,6 and 7-bit shuffles that do not align to computer bytes or words. To this end, we introduce (i) \emph{irregular product quantizers} combining sub-quantizers of different granularity and (ii) \emph{split tables} allowing lookup tables larger than registers. We evaluate Quicker ADC with multiple indexes including Inverted Multi-Indexes and IVF HNSW and show that it outperforms the reference optimized implementations (i.e., FAISS and polysemous codes) for numerous configurations. Finally, we release an open-source fork of FAISS enhanced with Quicker ADC.

	\end{abstract}

	}

\maketitle

\IEEEdisplaynontitleabstractindextext

\section{Introduction}
The Nearest Neighbor (NN) search problem consists in finding the closest vector $x$ to a query vector $y$ in a database of $N$ $d$-dimensional vectors. Efficient NN search in high-dimensional spaces is a core building-block in many multimedia retrieval applications, such as image similarity search, classification, or object recognition. These problems involve extracting high-dimensional feature vectors, or descriptors, and finding the NN of the extracted descriptors in a database of descriptors. For images, SIFT \cite{Lowe1999}, GIST~\cite{Oliva2001} or Deep-learning-based~\cite{Babenko2016} descriptors are often used.

Although efficient solutions for exact NN search have been proposed for low-dimensional spaces, \emph{exact} NN search remains challenging in high-dimensional spaces due to the notorious curse of dimensionality. Hence, much research work has been devoted to Approximate Nearest Neighbor (ANN) search. ANN search returns sufficiently close neighbors instead of the exact NN. Product Quantization (PQ) \cite{Jegou2011} is an ANN search used in numerous applications~\cite{Krapac2014,Xie2015}. PQ compresses high-dimensional vectors into short codes of a few bytes, enabling in-memory storage of large databases.

Fast answer is a key feature of PQ. It is enabled by Asymmetric Distance Computation (ADC), which efficiently computes distances between query vectors and compressed database vectors using in-memory lookup tables. Yet, despite being faster than regular distance computation, ADC remains bottlenecked by the many memory accesses it performs~\cite{Andre2015}. To date, much of the related work has been devoted to the development of efficient inverted indexes~\cite{Babenko2015, Xia2013}, which reduce the number of ADCs required to answer NN queries. Recently, there also has been a growing interest in increasing the performance of the ADC procedure itself with the introduction of PQ Fast Scan~\cite{Andre2015},  Quick ADC~\cite{Andre2017}, or polysemous codes~\cite{Douze2016}. Quick ADC leverages SIMD shuffle instructions to avoid memory accesses and implement very fast ADC; yet it is restricted to 4-bit sub-quantizers and has only been evaluated on simple inverted indexes, thus lacking performance results for more advanced indexes. Polysemous codes leverage the freedom to choose code indices to encode a binary code that is used to prune ADC computations using a low-cost hamming distance; yet, the use of an hamming distance affects the precision. 

In this paper, we present Quicker ADC, that introduces two new features to improve performance and accuracy through the use of the latest revision of SIMD instructions, namely AVX512: (i) \emph{irregular product quantizers} combining sub-quantizers of different sizes to allow using 5-bit or 6-bit sub-quantizers, (ii) \emph{split tables} for lookup tables larger than registers thus allowing efficient implementation of 8-bit sub-quantizer from 6-bit or 7-bit shuffles.
Quicker ADC is implemented into the reference library FAISS~\cite{FAISS,JDH17} to allow comparison to reference optimized implementations. We released it at \emph{\url{https://github.com/nlescoua/faiss-quickeradc}} in order to ease comparisons to our schemes, their adoption, and their evaluation in other settings. 

We compare the performance of Quicker ADC to PQ codes~\cite{Jegou2011} and polysemous codes~\cite{Douze2016} with multiple index types (i.e., simple inverted index~\cite{Jegou2011}, inverted multi indexes (IMI)~\cite{Babenko2015}, and inverted indexes based on HNSW~\cite{Malkov2016}) for both the SIFT1000M dataset and the Deep1B dataset. Quicker ADC consistently outperforms polysemous codes~\cite{Douze2016}, the state of the art solution for fast response time. For example, on SIFT1000M, for budget of 0.25ms per query and 128-bit codes, Quicker ADC (variant \ipq{24}{6,6,4}) achieves R@1 of 0.23 and R@100 of 0.60 with IMI ($K=4096^2$) and R@1 of 0.24 and R@100 of 0.68 with IVF HNSW ($K=2^{18}$), when polysemous codes achieve R@1 of 0.23 and R@100 of 0.47 with IMI and R@1 of 0.18 and R@100 of 0.36 with IVF HNSW.

\section{Background}
In this section, we first review the data-structures and algorithms used for nearest neighbor search with product quantization. We then discuss the impact of product-quantization parameters on search speed and recall. Finally, we introduce the capabilities of the latest processors supporting AVX-512 and their potential for supporting product quantization.

\subsection{Nearest neighbor search with PQ}

We describe the different steps for building an indexed database of vectors, and storing compressed representations of these vectors with product quantization. Such a database can then be used to search for the nearest neighbor of a \emph{query} vector efficiently by identifying a subset of \emph{database} vectors for which the distances to the \emph{query} vector needs to be computed, and computing efficiently these distances on compressed representations using a procedure called ADC.

\subsubsection{Index}

\label{sec:ivf}
Computing distances for all 1-billion vectors of a database is prohibitive. To tackle this issue in high-dimensional spaces, the common approach is to partition the database using a coarse indexing structure~\cite{Babenko2015,Jegou2011}. The input vector space is partitionned  into K Voronoi cells using a coarse-quantizer $q_i$. Vectors lying in each cell are stored in an inverted list. At query time, the inverted index is used to find the closest cells to the query vector, and distances for vectors in the inverted lists of these cells are computed. The two most popular approaches are inverted indexes (IVF)~\cite{Jegou2011,Jegou2011R} and multi-indexes~\cite{Babenko2015}. The number of cells is limited in IVF (e.g., $K=2^{16}$) due to the cost of training the IVF and computing distances to each cell. Multi-indexes~\cite{Babenko2015} push this limit and allow a more fine-grained index (e.g., $K=2^{24}$) at the price of imbalance in inverted list sizes. Recently, alternative approaches have leveraged nearest neighbor graphs to allow faster navigation in the index while avoiding the imbalance in inverted list sizes (e.g., HNSW)~\cite{Malkov2016,Baranchuk_2018_ECCV} for improved performance. Our work is orthogonal to these and compatible with any type of index. In addition, we will discuss in the evaluation the fact that Quicker ADC fits well with the latest indexes (e.g., HNSW). 

In order to fit the complete database into memory, short codes, which are much more compact, are stored instead of full vectors. To obtain a short code, the residual $\operatorname{r}(x) = x - \operatorname{q_i}(x) $ is encoded using a product quantizer described in the next section. Indexed databases therefore use two quantizers: a quantizer for the index ($\operatorname{q_i}$) and a product quantizer to encode residuals into short codes. The energy of residuals $\operatorname{r}(x)$ is smaller than the energy of input vectors $x$, thus there is a lower quantization error when encoding residuals rather than input vectors $x$ into short codes. In the rest of the paper, we will note $y = \operatorname{r}(x)$. As a special case, when product quantization  is used without any index, all vectors are stored in a single list. The short codes represent the input vectors rather than the residuals (i.e., $y=x$). 

\subsubsection{Short codes with product quantization}
\label{sec:vecenc}
\emph{Vector quantizers.} To encode residual vectors as short codes, PQ builds on vector quantizers. A vector quantizer $\operatorname{q}$ maps a vector $y \in \mathbb{R}^d$, to a vector $c \in \mathcal{C} \subset \mathbb{R}^d$. Vectors $c$ are called \emph{centroids}, and the set of centroids $\mathcal{C}$, of cardinality $k$, is the \emph{codebook}. Generally, the quantizer is chosen so that it maps the vector $y$ to its closest centroid $c$
\[
\operatorname{q}(y) = \argmin_{c \in \mathcal{C}}{||y-c||}
\]
The quantizer extends as an encoder $\operatorname{e}$ which encodes $y$ into the index $i \in \{0 \dots k-1\}$ of the vector $c_i \in C$ it is mapped onto (i.e., $\operatorname{e}(y) = i,\text{ such that }\operatorname{q}(y) = c_i$). The short code $i$ only occupies $b = \lceil \log_2(k) \rceil$ bits, which is typically much smaller than the $d\cdot32$ bits occupied by a vector $y \in \mathbb{R}^d$ stored as an array of $d$ single-precision floats (32 bit each). 

To maintain the quantization error sufficiently low for ANN search, a very large codebook \eg $k=2^{64}$ is required. However, building such codebooks is not tractable both in terms of processing and memory requirements.

\emph{Product quantizers.} Product quantizers overcome this issue by dividing a vector $y \in \mathbb{R}^d$ into $m$ sub-vectors, $y = (y^0,\dots,y^{m-1})$. Each sub-vector $y^j \in \mathbb{R}^{d/m}$, $j \in \{0,\dots,m-1\}$ is quantized independently using a sub-quantizer $\operatorname{q}^j$. Each sub-quantizer $\operatorname{q}^j$ has a distinct codebook $\mathcal{C}^j=(c_i^j)_{i=0}^{k-1}$ of cardinality $k$. 
The cardinality of the product quantizer codebook $\mathcal{C} = \mathcal{C}^0 \times \dots \times \mathcal{C}^{m-1}$ is $k^m$. Thus, a product quantizer has many centroids $k^m$
while only requiring storing and training $m$ codebooks of cardinality $k$. A product quantizer encodes a vector $y$ into a short code, by concatenating codes produced by sub-quantizers 
$\operatorname{e}(y) = (i_0\dots{}i_{m-1})$, such that $
\operatorname{q}(y) = \left ( \operatorname{q}^0(y^0)\dots\operatorname{q}^{m-1}(y^{m-1}) \right )
= \left (c^0_{i_0}\dots{}c^{m-1}_{i_{m-1}}\right )$. 
The code $(i_0,\dots,i_{m-1})$ requires $\lceil \log_2(k^m) \rceil$ bits of storage.

Interestingly, the order of vectors in the codebook $C^j$ is not constrained. Thus one can choose how vectors of the codebook are mapped to indexes. This freedom allows storing additional information and has been used in~\cite{Andre2015} to nest a 4-bit product quantizer into the 8-bit product quantizer and in~\cite{Douze2016} to encode a binary code onto the index. In both case, this allows pruning the computation thanks to the approximation of distance provided by the nested code. With binary code, the resulting combination is called \emph{polysemous codes}~\cite{Douze2016} and achieves state-of-the-art performance for approximate nearest neighbor search with product quantization (and inverted multi-indexes).

\begin{table*}[t]
	\footnotesize%
	\def\arraystretch{1.25}%
	\begin{threeparttable}%
		\caption{Instruction set capabilities}%
		\label{tbl:simdcap}%
		\begin{tabularx}{\linewidth}{|c|c|c|c|c|Y|Y|Y|}%
			\hline
			Instruction set & Instruction &Table size  & Distance & Lookups per operation & Latency, Rec. Througput 
			& Available since \\ 	\hline 
			SSE & \verb|pshufb|& \phantom{1}$16 = 2^4 \phantom{.\tnote{**}}$ & 8 bit & 16 & 1,1 & 2004 (Prescott) \\ 
			\hline
			AVX2 & \verb|pshufb| & \phantom{1}$16 = 2^4 \phantom{.\tnote{**}}$ & 8 bit & 32 & 1,1 &2013 (Haswell) \\ 
			\hline
			AVX512 BW & \verb|pshufb| & \phantom{1}$16 = 2^4 \phantom{.\tnote{**}}$& 8 bit & 64 & 1,1 & 2017 (Skylake SP) \\ 
			\hline
			AVX512 BW &\verb|vpermw| & \phantom{1}$32 = 2^5$ \tnote{*}\phantom{.} & 16 bit & 32 & 4,2 &2017 (Skylake SP) \\ 
			\hline
			AVX512 BW &\verb|vpermi2w|& \phantom{1}$64 = 2^6$ \tnote{**} & 16 bit & 32 & 7,2 &2017 (Skylake SP) \\ 
			\hline
			AVX512 VBMI &\verb|vpermb|& \phantom{1}$64 = 2^6\phantom{ \tnote{**}}$ & 8 bit & 64 & 3,1 & Exp. 2019 (Cannonlake) \\ 
			\hline
			AVX512 VMBI &\verb|vpermi2b| & \phantom{.}$128 = 2^7$ \tnote{**} & 8 bit & 64 & 5,2 & Exp. 2019 (Cannonlake) \\ 
			\hline 
			Neon &\verb|vtbl|& \phantom{1}$16 = 2^4 \phantom{.\tnote{**}}$ & 8 bit & 8 & 3,2~~~ (Cortex A53, A72) & 2009(ARMv7) \\ 
			\hline
			Neon &\verb|vtbl| & \phantom{.}$32 = 2^5 \phantom{.\tnote{**}}$ & 8 bit & 8 & 6,2~~~ (Cortex A53, A72)& 2009(ARMv7) \\
			\hline 
			VMX/AltiVec &\verb|vperm| & \phantom{.}$16 = 2^4 \phantom{.\tnote{**}}$ & 8 bit & 16 & Depend & 2006 (PowerISA v2.03)\color{black} \\
			\hline 
		\end{tabularx}
		
		\begin{tablenotes}
			
			\footnotesize
			\item \hspace{-1em}* Can be used for 16-values ($2^4$) table with 16-bit distances by zeroing the upper half of the table. \hfill ** Store the lookup table in two registers. 
		\end{tablenotes}
	\end{threeparttable}
	\def\arraystretch{1}%
	\normalsize
	\vspace{-0.5cm}
\end{table*}

\subsubsection{Search in the compressed domain}
Search for the nearest neighbor computes the distance between the query vector $z$ and a subset of database vectors. 

As a first step, the $a$ closest cells of the index quantizer $q_i$ are determined (typically, $a = 8$ to 64 for IVFs). The inverted lists of these cells correspond to all the candidate vectors for which the distances must be computed. For efficiency reasons, the distance is computed directly on the compressed representation using a procedure called ADC. 

ADC works the following way. First, for each cell, the residual $z' = \operatorname{r}(z)$ of the query vector is computed ($z' = z$ if no index is used). From this residual $z'$,  a set of $m$ lookup tables are computed $\{D^j\}_{j=0}^{m}$, where $m$ is the number of sub-quantizers of the product quantizer. The $j$th lookup table comprises the distance between the $j$th sub-vector of $z'$ and all centroids of the $j$th sub-quantizer:
\begin{equation}
D^j = \left( \left \lVert {z'}^j - \mathcal{C}^j[0] \right\rVert^2,\dots, \left \lVert {z'}^j - \mathcal{C}^j[k-1] \right \rVert^2\right) \label{eqn:tables}
\end{equation}
Second all candidates are scanned and the lookup tables are used to compute the distance between the query vector $z$ and each short code $c$ as follows:
\begin{equation}
\operatorname{adc}(z,c) = \sum_{j=0}^{m-1} D^{j}[c[j]] \label{eqn:adc1}
\end{equation}
Thus, ADC computes the distance between a query vector $z$ and each code $c$ by summing the distances between the sub-vectors of $z'$ and centroids associated with code $c$ in the $m$ sub-spaces of the product quantizer. As the number of codes in inverted lists is large compared to $k$, the number of centroids of sub-quantizers, using lookup tables avoids computing $\lVert {z'}^j - \mathcal{C}^j[i] \rVert^2$ for the same $i$ multiple times. Also, lookup tables provide a significant speedup by performing the computation directly in the compressed domain rather than reconstructing (i.e., decompressing) database vectors.

\subsubsection{Impact of PQ parameters}
\label{sec:paramimpact}

$m$, the number of sub-quantizers and $k$, the number of centroids of each sub-quantizer impact: (1) the memory usage of codes, (2) the recall of ANN search and (3) search speed.
The first tradeoff is between memory usage and accuracy. Both the memory usage of codes ($\lceil \log_2(k^m) \rceil$ bits = $m\cdot b$ bits, where $b=\lceil \log_2(k) \rceil$) and accuracy increase with the total number of centroids of the product quantizer ($k^m$). In practice, 64-bit or 128-bit codes are used in most cases.

The second tradeoff is between accuracy and search speed. For a constant memory budget of $m\cdot b$ bits per code, the respective values of $m$ and $b$ impact accuracy and speed. Decreasing $m$, which implies increasing $b$, increases accuracy~\cite{Jegou2011}. A detailled analysis of the impact of $m$ and $b$ on performance is given in\cite{Andre2015}. In a nutshell, $b$ impacts the time for each lookup in the table: if $b$ is too large, the lookup table does not fit into the fastest memory (i.e., the processor cache, which is limited in capacity) and lookup time will increase significantly. $m$ impacts the number of lookups: thus as long as the table stays in the fastest memory, the lower the $m$, the better the performance. In addition, the cost of computing the lookup tables grows exponentially with $b$, thus smaller $b$ also impact performance by reducing this cost; this is best seen on fine-grained indexes where the table computation cost becomes significant.

The standard notation for Product Quantization codes is $m\stimes{}b$ which specifies both parameters. The most common Product Quantization codes are $m\stimes{}8$ (e.g., \rpq{8}{8} or \rpq{16}{8}) as they ensure that tables fit in the processor cache and are not too costly to compute while being efficient to compute as they align well to computer bytes, and allow accessing the code without shifting nor masking.

\subsection{ADC computation using SIMD}
The common parameterizations of PQ (i.e., \rpq{m}{8} ) already exploit the fastest memory available on processors (i.e., the L1 cache), leaving no room for easy improvement.
A common technique to improve performance is to use instructions that process a vector of values rather than a single value at each CPU cycle: this principle called SIMD \emph{Single Instruction Multiple Data} allows significant performance boost for signal processing and matrix operations. Yet, for PQ, moving to SIMD improves performance for additions but the implementation remains bottlenecked by the memory accessed~\cite{Andre2015}. \begin{rev} In \cite{Andre2015}, the sequential implementation of ADC is analyzed thoroughly and compared to various approaches using in-memory lookup tables but with SIMD additions. The improvement is limited as the bottleneck is the access to memory and specifically the in-memory lookup tables as the codes, which are sequentially read, are efficiently handled by the hardware prefetchers of the processor.
\end{rev} Indeed SIMD does not allow an efficient implementation of in-memory table lookups, even using \texttt{gather} instructions introduced in recent processors~\cite{Andre2015,Hofmann2014}. While SIMD can add up to 16 floating-point numbers (512 bits) at once, only 2 concurrent memory accesses can be performed per cycle in each CPU core. Those memory accesses are the bottleneck. 

Thus, previous work~\cite{Andre2015,Andre2017,Blalock2017,Wu2017} moved lookup tables from memory to SIMD registers and leveraged in-register shuffles to implement lookups\footnote{Note that each core includes its own SIMD unit. We leave aside the use of multiple cores, as this is achieved easily by having multiple threads processing independent queries in parallel.}. Yet, the width of SIMD registers (128-512 bits) challenges this approach. Indeed, for common PQ (i.e., \rpq{m}{8} ), each lookup table occupies 8192 bits ($k=2^{8}=256$ floats). Previous work worked around this limit by (i) using 4-bit subquantizers and (ii) quantizing floats to 8-bit integers. The resulting lookup tables are thus small enough (128 bits) to fit SSE or AVX-2 registers. This has allowed Quick ADC~\cite{Andre2017} to achieve a significant performance improvement with a moderate loss of accuracy. This loss of accuracy comes mainly from the reduced precision of 4-bit subquantizers when compared to 8-bit subquantizers and to a smaller extent from the use of quantized distances. Yet, \cite{Andre2017,Blalock2017,Wu2017} remain limited to \rpq{m}{4} as they follow the initial approach of \cite{Andre2015} that use \texttt{pshufb}.

\subsection{SIMD capabilities and alternatives}
AVX512, introduced in 2017 with Xeon Scalable processors, is a significant redesign of Intel's SIMD instruction set. It provides additional shuffle instructions that support larger tables as described in Table~\ref{tbl:simdcap}. Available processors allow lookup tables of 32 or 64 16-bit values, thus 5-bit or 6-bit indexed lookup tables (i.e., $m\stimes 5$ or $m\stimes 6$ PQ codes). In addition, the wider registers (4$\stimes$ when compared to SSE and 2$\stimes$ when compared to AVX2) allow either an improved parallelism or more precise distances (16-bit instead of 8-bit). In the next section, we will explore how these new capabilities can be exploited for improving PQ efficiency. 

Interestingly, shuffle instruction \texttt{pshufb} (used in, e.g., $m\stimes 4$ PQ codes) can process between 128 (SSE) and 512 (AVX-512) bits of data per cycle while the \texttt{popcount} instruction (binary codes) can process only 64 bits of data per cycle. Hamming distance (used in, e.g., polysemous codes) is considered as much faster than product quantization's ADC (i.e., with in-memory lookup tables) thanks to these fast \texttt{popcount}. Yet,  \texttt{pshufb} is even faster\footnote{Mula et al.~\cite{Mula2018} also noticed that \texttt{popcount} is relatively inefficient when compared to vectorized instructions and thus designed algorithms to replace \texttt{popcount} by a few vectorized instructions including one or two \texttt{pshufb} for long-enough bit vectors. Yet, this does not allow any accuracy improvement contrary to Quicker ADC.} allowing product quantization to rival binary codes regarding distance computation speed. As a side note, our paper focuses on ADC, yet hamming distance is conceptually closer to SDC~\cite{Jegou2011}; SDC would trade accuracy for speed by avoiding costs associated to distance table computation. This could provide small product quantization codes competing directly with binary codes. 

While AVX512 has improved capabilities (larger tables) and increased parallelism (i.e., number of lookups per cycle) thanks to wider registers, the gain obtained may be partially cancelled by the fact that processor cores running AVX2 or AVX512 code are down-clocked and thus run at a lower frequency than processor cores running sequential or SSE code~\cite{AVX}. Thus, an experimental evaluation is needed to assess the potential of these instructions for PQ-based applications. This is even more salient for non-exhaustive search as the overall query time is the result of the index search, the distance table computation if any, and the ADC-based scanning of all candidates whose performance change differently as the code or index is altered. 

\begin{rev}
The work presented in this paper extends to ARM (Neon) and PowerPC (VMX) processors. Even if they have different instruction sets, they have similar limitations regarding the maximum number of concurrent memory accesses. They feature SIMD units with a 4-bit shuffle equivalent to \texttt{pshufb} and saturated arithmetics necessary to support Quicker ADC's \rpq{m}{4} codes. Beyond CPUs, the tradeoff between lookup tables in memory and registers should be taken into account when designing programs for GPUs or FPGAs because of fundamental hardware design tradeoffs between memory size and access speed/parallelism.  
\end{rev}


\section{Quicker ADC}

In this section, we describe Quicker ADC, a generalization of Quick ADC~\cite{Andre2017} that aims at improving accuracy through the use of the latest AVX-512 SIMD instructions. Interestingly, AVX-512 provides shuffles indexed by 5 or 6 bits, thus allowing a more precise quantization of vectors. Yet, their use for ADC is not straightforward as practical constraints prevent the use of $m\stimes 5$ or $m\stimes 6$ PQ codes. 

Product quantizers $m\stimes 8$ or $m\stimes 16$, that are commonly used, are composed of 8-bit or 16-bit sub-quantizers. This choice stems from the fact that manipulating byte-aligned or word-aligned values is both simpler and faster. Earlier work on using SIMD for PQ departs from these choices by relying on $m\stimes 4$ PQ codes yet benefits from the fact that 4-bit values naturally align to bytes as $2\stimes4 = 8$ bits. Unfortunately, $m\stimes 5$ or $m\stimes 6$ PQ codes rely on 5-bit or 6-bit values, neither of which align well to computer words (16 bits) or bytes (8 bits). This prevents a computationally efficient implementation of such PQ codes which is the purpose of using SIMD.

A naive approach could be to add padding by storing $3\stimes5$ bits of PQ codes in each 16-bit word and leaving one unused bit in each word. Our initial experiments with this approach on an exhaustive search in the SIFT1M dataset show that a $16\stimes{}4$ PQ (R@1\footnote{R@1 designates the recall of the closest vector when considering only the top vector returned by the algorithm. R@10 or R@100 are relaxed versions where the closest vector is considered as recalled if found in the top 10 vectors or top 100 vectors returned by the algorithm.} of 0.159) outperforms a $12\stimes{}5$ PQ codes (R@1 of 0.153). Our hypothesis is that using only 60 bits out of 64 bits outweights the benefits of using 5-bit subquantizers in place of 4-bit subquantizers.

To this end, we explore two solutions to avoid padding. The first solution, relying on new Irregular PQ codes, is presented in Section~\ref{sec:irpq}. The second solution combines multiple large shuffles to implement each lookup for 8-bit subquantizers; it anticipates the availability of AVX512 VBMI 7-bit shuffle in a near future. Both solutions require an SIMD-compatible memory layout similar to the one of Quick ADC~\cite{Andre2017} and described in Section~\ref{sec:memlayout}. Finally, Quicker ADC improves the distance quantization of Quick ADC~\cite{Andre2017} in order to allow using all 8 bits and not just 7 bits for improved accuracy as explained in Section~\ref{sec:distances}.

\subsection{Irregular PQ}
\label{sec:irpq}
As a first solution to alleviate the alignement issue, we propose Irregular Product Quantization which combines subquantizers of different sizes (4,5 and 6 bits) that can all be implemented in SIMD, such that their combination aligns well to words (16-bits). A first variant groups one 6-bit subquantizer with two 5-bit subquantizers for a total of 16-bits and a second variant groups two 6-bit subquantizers with one 4-bit subquantizers also for a total of 16 bits. Multiple such groups are combined to form the complete Product Quantizer. We will use the notation $m\stimes\{a,b,c\}$ for an Irregular Product Quantizer formed of $m$ sub-quantizers grouped by three sub-quantizers of $a$ bits, $b$ bits and $c$ bits. For example, the rest of the paper will often consider the 64-bit product quantizer $12\stimes\{6,6,4\}$ that has the following subquantizers $[6,6,4,\; 6,6,4,\; 6,6,4,\; 6,6,4]$. We note $g$ the number of sub-quantizers grouped (e.g., $g=3$ for \ipq{m}{6,6,4} and $g=2$ for \ipq{m}{4,4}); so that $m/g$ is the number of groups. 

With subquantizers of different precisions, the allocation of input dimensions to subquantizers cannot be uniform. With product quantizers~\cite{Jegou2011}, the input vector is split in sub-vectors of equal dimension which are then quantized independently. With irregular product quantizers, however, the input vector is not split in sub-vectors of equal dimensions so as to leverage the improved representation capabilities of finer subquantizers. We map input dimensions proportionally to the bit-width of the subquantizer. Let us consider a 128-dimension SIFT vector quantized by an irregular product quantizer $12\stimes\{6,6,4\}$. Each of the 4 groups of sub-quantizers is associated to 32 dimensions. Thus within each group, 6-bit sub-quantizers quantize sub-vectors of 12 dimensions and 4-bit subquantizers quantize sub-vectors of 8 dimensions. Note that as long as $m/g$ is a multiple of 2, Irregular Product Quantizers remain compatible with multi-indexes which requires the two subquantizers of the index to be aligned with the subquantizers of the product quantizer.

If the number of input dimension cannot be divided exactly, the remaining dimensions are added one by one to the subquantizers. Yet, this alters performance and should be avoided: one should always ensure that allocation can be proportional. More specifically, when vectors are pre-processed by a PCA or a rotation (like in OPQ), the pre-processed vectors should be of a dimension which can be divided exactly. This implies that for a $m\stimes\{6,6,4\}$, the number of dimensions of the pre-processed vectors must be divisible by $(3+3+2)m/3 = 8m/3$, and for a $m\stimes\{6,5,5\}$ the number of dimension of the pre-processed vector must be divisible by $(6+5+5)m/3 = 16m/3$. For example, 128-dimension SIFT vectors can be encoded optimally by both $12\stimes\{6,6,4\}$ or $12\stimes\{6,5,5\}$ irregular PQ codes.

To validate this first solution, we compare the different 64-bit codes using an exhaustive search in the SIFT1M dataset. Both $12\stimes\{6,6,4\}$ (R@1 of 0.179) and $12\stimes\{6,5,5\}$ (R@1 of 0.174) outperform a $16\stimes 4$ PQ code (R@1 of 0.158). Additional results are given in the evaluation in Table~\ref{tab:exhaust}.

\begin{figure}
	\centering
	\begin{tikzpicture}
\footnotesize
\node[register,fill=black!5] (indexes01)
{%
  \nodepart[eltb,align=right,fill=black!10]{one}$a_0$
  \nodepart[eltb,align=right]{two}$b_0$
  \nodepart[eltb,align=right]{three}$c_0$
  \nodepart[eltb,align=center]{four}$\dots$%
  \nodepart[eltb,align=right]{five}$z_0$%
};

\node[above=0.1*\vertSep of indexes01.north] {loaded (compressed) vectors components};

\node[register, below=3*\vertSep of indexes01.south] (masked01)
{%
	\nodepart[eltb,align=left]{one}$a_0^8$
\nodepart[eltb,align=left]{two}$b_0^8$
\nodepart[eltb,align=left]{three}$c_0^8$
\nodepart[eltb,align=center]{four}$\dots$%
\nodepart[eltb,align=left]{five}$z_0^8$%
};

\node[register, below=3*\vertSep of masked01.south] (dist1a)
{%
	\nodepart[eltb]{one}$D^0[0]$
	\nodepart[eltb]{two}$D^0[1]$
	\nodepart[eltb]{three}$D^0[2]$
	\nodepart[eltb]{four}$\dots$%
	\nodepart[eltb]{five}$D^0[64]$%
};

\node[register, below=1*\vertSep of dist1a.south] (dist1b)
{%
	\nodepart[eltb]{one}$D^0[65]$
	\nodepart[eltb]{two}$D^0[66]$
	\nodepart[eltb]{three}$D^0[67]$
	\nodepart[eltb]{four}$\dots$%
	\nodepart[eltb]{five}$D^0[128]$%
};

\node[register, below=1.5*\vertSep of dist1b.south] (dist1c)
{%
	\nodepart[eltb]{one}$D^1[0]$
	\nodepart[eltb]{two}$D^1[1]$
	\nodepart[eltb]{three}$D^1[2]$
	\nodepart[eltb]{four}$\dots$%
	\nodepart[eltb]{five}$D^1[64]$%
};

\node[register, below=1*\vertSep of dist1c.south] (dist1d)
{%
	\nodepart[eltb]{one}$D^1[65]$
	\nodepart[eltb]{two}$D^1[66]$
	\nodepart[eltb]{three}$D^1[67]$
	\nodepart[eltb]{four}$\dots$%
	\nodepart[eltb]{five}$D^1[128]$%
};

\draw[->,thick] (indexes01.west) -- ++(-1*\horSep,0) |- (masked01.178);

\node[left= 0.2cm of masked01.south west,rotate=90,anchor=south west]{maskmov};

\def\r{{1..7}}

\node[register, below=3*\vertSep of dist1d.south] (part1a)
{%
	\nodepart[eltb]{one}$D^0[a^\r_0]$
	\nodepart[eltb]{two}$D^0[b^\r_0]$
	\nodepart[eltb]{three}$D^0[c^\r_0]$
	\nodepart[eltb]{four}$\dots$%
	\nodepart[eltb]{five}$D^0[z^\r_0]$%
};

\node[register, below=\vertSep of part1a.south] (part1b)
{%
	\nodepart[eltb]{one}$D^1[a^\r_0]$
\nodepart[eltb]{two}$D^1[b^\r_0]$
\nodepart[eltb]{three}$D^1[c^\r_0]$
\nodepart[eltb]{four}$\dots$%
\nodepart[eltb]{five}$D^1[z^\r_0]$%
};

\node[register, below=\vertSep of part1b.south,fill=black!10] (res)
{%
	\nodepart[eltb]{one}$\!\!D^{a^8_0}[a^\r_0]$
	\nodepart[eltb]{two}$\!\!D^{b^8_0}[b^\r_0]$
	\nodepart[eltb]{three}$\!\!D^{c^8_0}[c^\r_0]$
	\nodepart[eltb]{four}$\dots$%
	\nodepart[eltb]{five}$\!\!D^{z^8_0}[z^\r_0]$%
};

\draw[->,thick] (indexes01.east) -- ++(1*\horSep,0) |- (part1a.east);
\draw[->,thick] (dist1a.east) -- ++(1*\horSep,0) |- (part1a.east);
\draw[->,thick] (dist1b.east) -- ++(1*\horSep,0) |- (part1a.east);

\draw[->,black!30,thick] (indexes01.east) -- ++(3*\horSep,0) |- (part1b.east);
\draw[->,black!30,thick] (dist1c.east) -- ++(3*\horSep,0) |- (part1b.east);
\draw[->,black!30,thick] (dist1d.east) -- ++(3*\horSep,0) |- (part1b.east);

\coordinate[below=0.58cm of res.south west]  (y1a) ;
\coordinate[below=0.75cm of res.two]  (y2a) ;
\draw [<->] (y1a) -- (y2a) node[midway,below=0.00cm] {\small 8 bits};

\coordinate[below=0.49cm of res.south west]  (y1b) ;
\coordinate[below=0.49cm of res.south east]  (y2b) ;
\draw [<->] (y1b) -- (y2b) node[pos=0.67,below=0.00cm] {\small 512 bits \hspace{1cm}  (j=0, g=1)};


\node[right= 0.5cm of dist1a.north east,rotate=270,anchor=north]{lookup (shuffle)};

\draw[-,dashed] (masked01.182) -- ++(-1*\horSep,0) |- (res.west);

\draw[->,thick] (part1a.west) --  ++(-1*\horSep,0) |- (res.west);
\draw[->,thick] (part1b.west) --  ++(-1*\horSep,0) |- (res.west);

\node[left= 0.2cm of part1b.north west,rotate=90,anchor=south]{blend};

\node[below=0.1*\vertSep of res.south] {partial distances for the loaded components};

\end{tikzpicture}
	\vspace{-0.3cm}
	\caption{Shuffle-blend lookup for $8\stimes\{8\}$ (AVX-512 VBMI)}
	\label{fig:lookup_blend}
\end{figure}
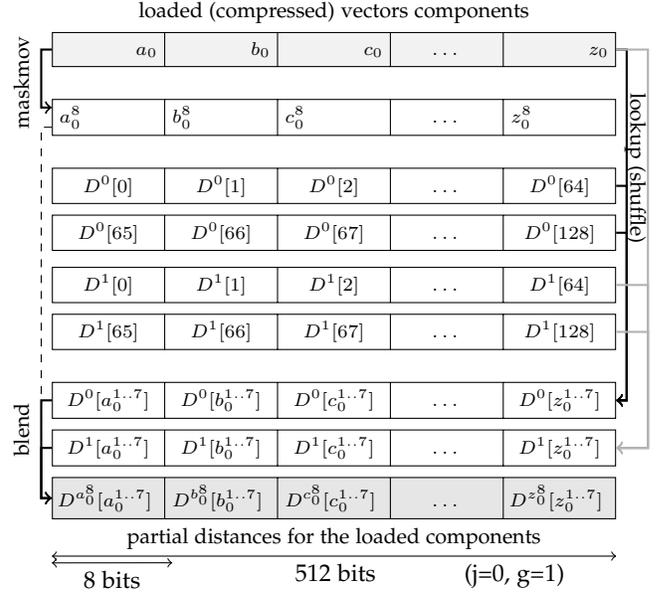

\setlength\tabcolsep{4 pt}
\begin{table}
		\caption{SIMD operations required to perform 8-bit lookups}%
		\label{tab:count}%
	\footnotesize%
	\centering%
	\begin{tabular}{|c|c|c|c|c|c|}%
		\hline 
		& Native & \multicolumn{2}{c|}{Op. count}  & Values per  & Ops per \\
		Instruction set & Shuffle & Shuffle & Blend & register  & value \\ 
		\hline 
		AVX2 / SSE4 & 4-bit & 16 & 15 & 32 / 16 & 0.97 / 1.94 \\ 
		\hline  
		AVX512 BW & 6-bit & 4 & 3  & 32 & 0.21\\ 
		\hline 
		AVX512 VBMI & 7-bit & 2 & 1 & 64 & 0.05\\ 
		\hline 
	\end{tabular} 

\end{table}

\begin{figure*}
	\centering%
	\subfloat[Standard layout \protect\\($m\stimes{}8$ or $m\stimes{}16$)]{%
		\centering%
		\hspace{0.2cm}
		\vbox{
			\hbox{\begin{tikzpicture}
  \footnotesize
\node [memorys,
  column 1/.style={text=black!70},
] (layout)
{ a_0 & \dots & a_{m-1} \\
  b_0 & \dots & b_{m-1} \\
  \dots & \dots & \dots \\
  z_0 & \dots & z_{m-1} \\
};

  \node[label, left=\horSep of layout-1-1.west, anchor=east] {$a$};
  \node[label, left=\horSep of layout-2-1.west, anchor=east] {$b$};
  \node[label, left=\horSep of layout-4-1.west, anchor=east] {$z$};

\end{tikzpicture}}%
			\vspace{0.05cm}
		}
		\hspace{0.2cm}
		\label{fig:layoutnorm}%
	}
	\hfill
	\subfloat[Transposed layout ($m\stimes\{4,4\}$)\protect\\QuickerADC with pshufb, 8-bit distances]{%
		\centering%
		  \begin{tikzpicture}
  \footnotesize
  \node [memory, row 1/.style={text=black!70}] (layout)
  { a_1\,a_0  &\dots & p_1\,p_0 \\
    \dots  & \dots  & \dots  \\
    a_{m-1}\,a_{m-2} & \dots & p_{m-1}\,p_{m-2} \\
  };
  
  \node[label, above=\vertSep of layout-1-1.north, anchor=south] {$a$};
  \node[label, above=\vertSep of layout-1-2.north, anchor=south] {$b$};
\node[label, above=\vertSep of layout-1-3.north, anchor=south] {$p$};

\coordinate[below=0.18cm of layout-3-1.south west]  (y1a) ;
\coordinate[below=0.18cm of layout-3-1.south east]  (y2a) ;
\draw [<->] (y1a) -- (y2a) node[midway,below=0.00cm] {  \footnotesize 8 bits};

\coordinate[below=0.09cm of layout-3-1.south west]  (y1b) ;
\coordinate[below=0.09cm of layout-3-3.south east]  (y2b) ;
\draw [<->] (y1b) -- (y2b) node[pos=0.57,below=0.00cm] {  \footnotesize 128 bits};


\coordinate[left=0.2cm of layout-1-1.north west]  (y1) ;
\coordinate[left=0.2cm of layout-3-1.south west]  (y2) ;
\draw [<->] (y1) -- (y2) node[midway,rotate=90,above] {  \scriptsize $m/2$ groups};
  \end{tikzpicture}%
		\label{fig:layouttrans}%
	}
	\hfill
	\subfloat[Transposed layout ($m\stimes\{6,6,4\})$\protect\\QuickerADC with vpermw/i2w, 16-bit distances]{%
		\centering%
		  \begin{tikzpicture}
    \footnotesize
  \node [memory16, row 1/.style={text=black!70}, col 2/.style={nodes=memorys}] (layout)
  { a_2\,a_1\,a_0  & \dots & z_2\,z_1\,z_0 \\
    \dots & \dots & \dots  \\
    a_{m-1}\,a_{m-2}\,a_{m-3} & \dots & z_{m-1}\,z_{m-2}\,z_{m-3} \\
  };
  
  \node[label, above=\vertSep of layout-1-1.north, anchor=south] {$a$};
  \node[label, above=\vertSep of layout-1-2.north, anchor=south] {$b$};
\node[label, above=\vertSep of layout-1-3.north, anchor=south] {$z$};

\coordinate[below=0.18cm of layout-3-1.south west]  (y1a) ;
\coordinate[below=0.18cm of layout-3-1.south east]  (y2a) ;
\draw [<->] (y1a) -- (y2a) node[midway,below=0.00cm] {\footnotesize 16 bits};

\coordinate[below=0.09cm of layout-3-1.south west]  (y1b) ;
\coordinate[below=0.09cm of layout-3-3.south east]  (y2b) ;
\draw [<->] (y1b) -- (y2b) node[pos=0.57,below=0.00cm] {\footnotesize 512 bits};

\coordinate[left=0.2cm of layout-1-1.north west]  (y1) ;
\coordinate[left=0.2cm of layout-3-1.south west]  (y2) ;
\draw [<->] (y1) -- (y2) node[midway,rotate=90,above] {\scriptsize $m/3$ groups};

  \end{tikzpicture}%
		\label{fig:layouttrans_512}%
	} 
	\caption{Inverted list memory layouts. Each table cell represents a byte or a word.\label{fig:layout}}
\end{figure*}
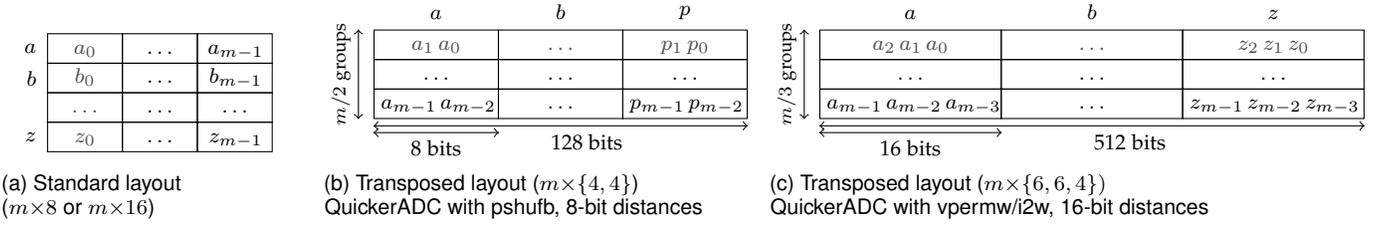

\subsection{Split tables}
\label{sec:blend}
AVX-512 brings 6-bit \texttt{vpermi2w} and 7-bit \texttt{vpermi2b} shuffles. These are only 2-bit away and 1-bit away from the very common 8-bit sub-quantizers. Rather than relying on highly-imbalanced irregular product quantizers, in particular a $m\stimes\{7,1\}$ that would perform poorly, it becomes interesting to consider that each 8-bit lookup table can be split into 4 6-bit lookup tables or 2 7-bit lookup tables. As an example, we show an 8-bit lookup built from two 7-bit shuffles on Figure~\ref{fig:lookup_blend}. The distance table $D$ is split in two halves ($D^0$ and $D^1$, each of which occupies 2 registers). Two shuffles (each having 3 registers as inputs) are performed to get values indexed by the low 7-bits (e.g., $a^{1..7}_0)$). The final values are selected through a blend indexed by the 8-th bit (e.g., $a^{8}_0$).

This approach cannot be built on 4-bit shuffle (\texttt{pshufb} from SSE4/AVX2) as it would require too many shuffles and blends. As shown in Table~\ref{tab:count}, performing an 8-bit lookup for 16 values using SSE4 requires 16 shuffles and 15 blends, an average of 1.94 operations/value. In comparison, when using AVX512 VBMI, only 0.05 operations/value are necessary. Due to the lack of available processors with AVX512 VBMI, we will evaluate performance only for AVX512 BW (0.21 operations/value); yet, it is clear that \texttt{vpermi2b} from AVX512 VBMI will provide significant gains\footnote{The software we release already includes an implementation supporting AVX512 VBMI even if it couldn't be evaluated yet.}.

Quicker ADC codes implemented with this approach require more instructions per lookup than irregular product quantizers, yet, they come with almost no compromises on accuracy when compared to \pq{m}{8} PQ codes. In the evaluation, we'll see that in some context they outperform the alternative approach already, but they will become particularly interesting as AVX512 VBMI-capable processors become available in the near future.

\subsection{Memory layout}
\label{sec:memlayout}
Similarly to Quick ADC, Quicker ADC requires a transposed memory layout. 
Indeed, an SIMD in-register shuffle performs multiple lookups at once, but in a \emph{single} lookup table \eg $D^0$. Therefore, shuffles must operate on a single component of multiple codes (e.g., $a_0,\dotsc,p_0$) at once, and not on multiple components of a single code (e.g., $a_{0},\dotsc,a_{15}$). Hence, to allow efficient loads from memory, all values of the SIMD register $a_0,\dotsc,p_0$ must be \emph{contiguous} in memory, which is not the case with the memory layout of inverted lists (Figure \ref{fig:layoutnorm}) used in common PQ implementations~\cite{Jegou2011,Douze2016}.

The size of the blocks transposed depends on the shuffle used. With 4-bit shuffles (\texttt{pshufb} in SSE4/AVX2/AVX512) operating on lanes of 128 bits composed of 16 values, we transpose blocks of 16 codes ($a \text{--}p$) as shown on Figure~\ref{fig:layouttrans}. With 5-bit and 6-bit shuffles (\texttt{vpermw} and \texttt{vpermi2w} in AVX512) operating on a single lane of 512 bits composed of 32 values, we transpose blocks of 32 codes ($a\text{--}z$) as shown on Figure~\ref{fig:layouttrans_512}. With 7-bit shuffles (\texttt{vpermi2b} from AVX512) operating on 64 values, we transpose blocks of 64 codes. \trev{A more straightforward approach would transpose to blocks of the size of a register (rather than the size of a lane), but this would hinder performance due to increased register pressure. Also for smaller sets of vectors encountered in inverted indexes, this would require additional padding resulting in lower performance.}

In addition, values have a fixed width that depends on the shuffle instruction used: \texttt{pshufb} and \texttt{vpermb} operate on bytes (8 bits) while \texttt{vpermw} and \texttt{vpermi2w} operate on words (16 bits). Hence, multiple subcodes must be packed together to form bytes or words. We use the notation of irregular product quantizers to specify the packing applied (e.g.,\ipq{m}{6,6,4} packs 3 subcodes to form a 16-bit word ($6+6+4=16$)  as shown on Figure~\ref{fig:layouttrans_512}). This notation extends to regular product quantizers (e.g., Quick ADC~\cite{Andre2017}'s \ipq{m}{4,4} packs 2 4-bit subcodes to form an 8-bit byte ($4+4=8$) as shown on Figure~\ref{fig:layouttrans}). Split-table-based \ipq{m}{8,8} packs two 8-bit subcodes to form a word for use with \texttt{vpermi2w}, and split-table-based \ipq{m}{8} packs a single 8-bit subcode to form a byte for use with \texttt{vpermi2b}. Subcodes can be extracted from the packed values efficiently in SIMD through shift and mask. Note that the number of rows in the transposed layout on Figure~\ref{fig:layout} corresponds exactly to the number of groups in irregular PQ codes.

\subsection{Quantization of distances}
\label{sec:distances}

In standard ADC, lookup tables store partial distances as 32-bit floats. As subquantizer precision is key to accuracy, we seek to store partial distances as 8-bit or 16-bit integers so as to allow lookup tables of the same size, yet storing twice or four times as much values. The representation (8-bit or 16-bit integers) depends on whether the type of shuffle we use operates on bytes (e.g., \texttt{pshufb}, \texttt{vpermi2b}) or words (e.g, \texttt{vpermw}, \texttt{vpermi2w}).

As we are interested only in the top-$k$ nearest neighbors, our distance quantization scheme must represent as precisely as possible the smallest distances, but can ignore (i.e., quantize to $\infty$) large distances. Thus, to perform distance computations (additions, ...), we rely on saturated integer arithmetics that handles $\infty$ through saturation. The approach is thus similar to that of Quick ADC but provides a tighter distance quantization as explained hereafter. 

\trev{First, we use 8-bit and 16-bit \emph{unsigned} integers whereas Quick ADC uses only the 7-bit positive range of \emph{signed} integers: this doubles the precision. Note that AVX2 lacks straigtforward instructions for unsigned comparisons~\cite{Intrisics}.}

Second, we perform a tighter evaluation of the minimum and maximum values than in Quick ADC~\cite{Andre2017} in order to allow a more precise quantization. For each of the $m$ lookup tables, we evaluate the smallest partial distance $p_\textrm{min}(i)$ to represent, which is the smallest partial distance in the $i$-th table. This also gives us the smallest distance to represent $d_\textrm{min} = \sum_{i=0}^m {p_\textrm{min}(i)}$. Then, we scan $t$ vectors to find a candidate set of $R$ nearest neighbor candidates, where $R$ is the number of nearest neighbors requested by the user (e.g., R=100 when R@100 is evaluated) and $t$ is larger than $R$ but much smaller than the total number of vectors. We use the distance of the query vector to the $R$-th nearest neighbor candidate \ie the farthest candidate, as the $d_\textrm{max}$ bound. All subsequent candidates have to be closer to the query vector, thus $d_\textrm{max}$ is the maximum distance we need to represent.

We determine the size of quantization bins $\Delta = \frac{d_\textrm{max} - d_{\textrm{min}}}{q_\textrm{max}}$. Partial distance $p$ in the $i$-th table can thus be quantized as%
	$$%
    q = \frac{p-p_\textrm{min}(i)}{\Delta}%
    $$%
To unquantize the sum $\Sigma{}q$, one can use:%
    $$%
    d = (\Sigma{}q)\Delta + d_\textrm{min}%
    $$%
Note that similarly to \cite{Andre2015,Andre2017}, we learn our distance quantizer at query time \trev{on summed distances from the top-$k$} rather than at training time \trev{on partial distances (i.e., values in the distance tables)~\cite{Blalock2017}}: the required evaluation of a few distances has a negligible impact on performance yet allows consistently increased accuracy. \trev{Also, contrary to~\cite{Blalock2017}, we limit distortion only for the shortest distances, whereas in~\cite{Blalock2017}, the distortion is limited for all partial distances, but all the shortest distances (i.e., below the first quantile) are quantized into a single bin resulting in higher distortion for these. Finally, as our $q_\texttt{max}$ bound is determined from summed distances, we can perform the accumulation on the same integer width as the lookups (i.e., no need to upcast 8-bit distances to 16-bit to avoid saturation) and preserve the semantic of $q_\texttt{max}$ (i.e., too large distance) during accumulation.}


\subsection{SIMD distance computation}
Quicker ADC supports several combinations of sub-quantizers: (i) those operating on 128-bit lanes for 4-bit subquantizers with 8-bit distances (SSE, AVX2 and AVX512),  (ii) those operating on 512 bits lanes for 4,5 and 6-bit subquantizers with 16-bit distances, and (iii) those operating on 512 bits lanes for 8-bit subquantizers with 8 or 16-bit distances implemented using multiple shuffles and blends as described in Section~\ref{sec:blend} and Figure~\ref{fig:lookup_blend}. While implementation details vary, the overall principle is the same.

\begin{figure}
	\centering
	\scalebox{0.91}{\begin{tikzpicture}
\footnotesize
\node[register,fill=black!5] (indexes01)
{%
  \nodepart[elt,align=right,fill=black!10]{one}$a_2\,a_1\,a_0$
\nodepart[elt,align=right]{two}$b_2\,b_1\,b_0$
\nodepart[elt,align=right]{three}$c2\,c_1\,c_0$
\nodepart[elt,align=center]{four}$\dots$%
\nodepart[elt,align=right]{five}$z_2\,z_1\,z_0$%
};

\node[above=0.1*\vertSep of indexes01.north] {loaded (compressed) $g$ first vectors components};

\node[register, below=3*\vertSep of indexes01.south] (masked01)
{%
	\nodepart[elt,align=right]{one}$a_0$
	\nodepart[elt,align=right]{two}$b_0$
	\nodepart[elt,align=right]{three}$c_0$
	\nodepart[elt,align=center]{four}$\dots$%
	\nodepart[elt,align=right]{five}$z_0$%
};

\node[register, below=\vertSep of masked01.south] (masked02)
{%
	\nodepart[elt,align=right]{one}$a_1$
\nodepart[elt,align=right]{two}$b_1$
\nodepart[elt,align=right]{three}$c_1$
\nodepart[elt,align=center]{four}$\dots$%
\nodepart[elt,align=right]{five}$z_1$%
};

\node[register, below=\vertSep of masked02.south] (masked03)
{%
	\nodepart[elt,align=right]{one}$a_2$
\nodepart[elt,align=right]{two}$b_2$
\nodepart[elt,align=right]{three}$c_2$
\nodepart[elt,align=right]{four}$\dots$%
\nodepart[elt,align=right]{five}$z_2$%
};

\node[register, below=3*\vertSep of masked03.south] (dist1a)
{%
	\nodepart[elt]{one}$D^0[0]$
	\nodepart[elt]{two}$D^0[1]$
	\nodepart[elt]{three}$D^0[2]$
	\nodepart[elt]{four}$\dots$%
	\nodepart[elt]{five}$D^0[31]$%
};

\node[register, below=\vertSep of dist1a.south] (dist1b)
{%
	\nodepart[elt]{one}$D^0[32]$
\nodepart[elt]{two}$D^0[33]$
\nodepart[elt]{three}$D^0[34]$
\nodepart[elt]{four}$\dots$%
\nodepart[elt]{five}$D^0[63]$%
};

\node[register, below=2*\vertSep of dist1b.south] (dist1c)
{%
	\nodepart[elt]{one}$D^1[0]$
	\nodepart[elt]{two}$D^1[1]$
	\nodepart[elt]{three}$D^1[2]$
	\nodepart[elt]{four}$\dots$%
	\nodepart[elt]{five}$D^1[31]$%
};

\node[register, below=2*\vertSep of dist1c.south] (dist1d)
{%
	\nodepart[elt]{one}$D^2[0]$
	\nodepart[elt]{two}$D^2[1]$
	\nodepart[elt]{three}$D^2[2]$
	\nodepart[elt]{four}$\dots$%
	\nodepart[elt]{five}$0$%
};

\draw[->,thick] (indexes01.west) -- ++(-1*\horSep,0) |- (masked01.west);
\draw[->,thick] (indexes01.west) -- ++(-1*\horSep,0) |- (masked02.west);
\draw[->,thick] (indexes01.west) -- ++(-1*\horSep,0) |- (masked03.west);

\node[left= 0.2cm of masked02.north west,rotate=90,anchor=south]{\footnotesize{} shift and mask};

\node[register, below=3*\vertSep of dist1d.south] (part1a)
{%
	\nodepart[elt]{one}$D^0[a_0]$
	\nodepart[elt]{two}$D^0[b_0]$
	\nodepart[elt]{three}$D^0[c_0]$
	\nodepart[elt]{four}$\dots$%
	\nodepart[elt]{five}$D^0[z_0]$%
};

\node[register, below=\vertSep of part1a.south] (part1b)
{%
	\nodepart[elt]{one}$D^1[a_1]$
	\nodepart[elt]{two}$D^1[b_1]$
	\nodepart[elt]{three}$D^1[c_1]$
	\nodepart[elt]{four}$\dots$%
	\nodepart[elt]{five}$D^1[z_1]$%
};

\node[register, below=\vertSep of part1b.south] (part1c)
{%
	\nodepart[elt]{one}$D^2[a_2]$
	\nodepart[elt]{two}$D^2[b_2]$
	\nodepart[elt]{three}$D^2[c_2]$
	\nodepart[elt]{four}$\dots$%
	\nodepart[elt]{five}$D^2[z_2]$%
};

\node[register, below=\vertSep of part1c.south,fill=black!10] (res)
{%
	\nodepart[elt]{one}$acc_a$
	\nodepart[elt]{two}$acc_b$
	\nodepart[elt]{three}$acc_c$
	\nodepart[elt]{four}$\dots$%
	\nodepart[elt]{five}$acc_z$%
};

\draw[->,thick] (masked01.east) -- ++(1*\horSep,0) |- (part1a.east);
\draw[->,thick] (dist1a.east) -- ++(1*\horSep,0) |- (part1a.east);
\draw[->,thick] (dist1b.east) -- ++(1*\horSep,0) |- (part1a.east);

\draw[->,black!30,thick] (masked02.east) -- ++(3*\horSep,0) |- (part1b.east);
\draw[->,black!30,thick] (dist1c.east) -- ++(3*\horSep,0) |- (part1b.east);

\draw[->,black!60,thick] (masked03.east) -- ++(5*\horSep,0) |- (part1c.east);
\draw[->,black!60,thick] (dist1d.east) -- ++(5*\horSep,0) |- (part1c.east);

\node[right= 1.2cm of dist1d.north east,rotate=270,anchor=north]{\footnotesize lookup (shuffle)};

\draw[->,thick] (part1a.west) --  ++(-1*\horSep,0) |- (res.west);
\draw[->,thick] (part1b.west) --  ++(-1*\horSep,0) |- (res.west);
\draw[->,thick] (part1c.west) --  ++(-1*\horSep,0) |- (res.west);

\node[left= 0.2cm of part1c.north west,rotate=90,anchor=south]{add};
\node[below=0.1*\vertSep of res.south] {partial distances for $g$ first components};

\coordinate[below=0.58cm of res.south west]  (y1a) ;
\coordinate[below=0.775cm of res.two]  (y2a) ;
\draw [<->] (y1a) -- (y2a) node[midway,below=0.00cm] {\footnotesize 16 bits};

\coordinate[below=0.49cm of res.south west]  (y1b) ;
\coordinate[below=0.49cm of res.south east]  (y2b) ;
\draw [<->] (y1b) -- (y2b) node[pos=0.67,below=0.00cm] {\footnotesize 512 bits \hspace{1cm}  (j=0, g=3)};

\end{tikzpicture}}
	\caption{SIMD Lookup-add in AVX-512 for $12\stimes\{6,5,4\}$ \label{fig:lookup_add}}
\end{figure}
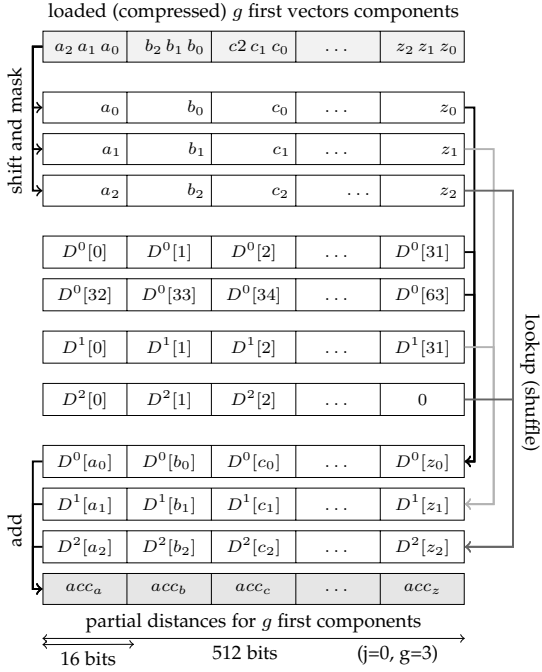

Once cells are selected and distance tables are computed, as explained in the background, each invert list is scanned block by block. The distances for vectors of the block are computed in the following way. We depict the processing applied to each group of components (i.e., each row of Figure~\ref{fig:layouttrans} and ~\ref{fig:layouttrans_512}) in Figure~\ref{fig:lookup_add}. First, the subcodes are unpacked using shifts and masks. For each set of subcodes, the partial distances are looked up in the distance table using either a native shuffle, as explained in Section~\ref{sec:irpq} or a lookup implemented through a combination of shuffle and blends, as explained in Section~\ref{sec:blend} and depicted in Figure~\ref{fig:lookup_blend}. The distances are summed using saturated arithmetic. Note that distance tables may occupy half, one or multiple registers depending on the type of lookup. Figure~\ref{fig:lookup_add} presents an hypothetical \ipq{12}{6,5,4} code in which distance tables for 6-bit, 5-bit and 4-bit subquantizers fit in respectively 2, 1 and half a register. This process is repeated for all $m/g$ groups that form the complete code and partial distances are summed to obtain the distances. The distances are compared to the worst candidate vector, and vectors for which distances are smaller are added to the binary heap of candidates.%
\newcommand{\simd}[1]{{\ttfamily\textls[-65]{#1}}}%
\newcommand{\dtype}[1]{{\ttfamily\textls[-65]{#1}}}%
\newcommand{\result}[3]{#1/#2\hfill\textit{#3}}%
\newcolumntype{?}{!{\vrule width 1.5pt}}%
\def\b#1{\textbf{#1}}%
\newcommand{\hfilll}{\hskip 0pt plus 1filll}%
\newcommand{\sspace}{\hspace{0.7em}}%
\begin{table*}%
		\caption{Exhaustive search (without index) on SIFT1M and Deep1M. Results are given as percentage\xspace\xspace R@1/R@100\xspace\xspace \textit{time(ms)}\vspace{-2em}}%
		\label{tab:exhaust}%
		\scalebox{0}{%
			\begin{tikzpicture}%
			\begin{axis}[]%
			\addplot[only marks, mark=diamond,draw=blue!70!black,line width=0.75pt,mark size=2.5pt] (0,0); \label{llpq}%
			\addplot[only marks, mark=square,draw=red!70!black,line width=0.75pt,mark size=2.5pt] (0,0); \label{llpoly}%
			\addplot[only marks, mark=triangle,draw=green!70!black,line width=0.75pt,mark size=2.5pt] (0,0);\label{ll44}%
			\addplot[only marks, mark=o,draw=white!50!black,line width=0.75pt,mark size=2.5pt] (0,0); \label{ll664}%
			\addplot[only marks, mark=star,draw=white!70!black,line width=0.75pt,mark size=2.5pt] (0,0); \label{ll655}%
			\addplot[only marks, mark=x,draw=orange,line width=0.75pt,mark size=2.5pt] (0,0); \label{ll88}%
			\addplot[only marks, mark=Mercedes star,draw=black,line width=0.75pt,mark size=2.5pt] (0,0); \label{llu}%
			\end{axis}%
			\end{tikzpicture}%
		}%

\colorbox{white}{%
\begin{threeparttable}%
\scriptsize%
	\begin{tabularx}{1.0\linewidth}{|l c| c | c ? Y|Y|Y ? Y|Y|Y|}%
		\hline \multicolumn{4}{|c?}{Results as: \xspace\xspace\xspace R@1/R@100(\%) \xspace\xspace\xspace\xspace\xspace\xspace\textit{time(ms)}}
		& \multicolumn{3}{c?}{\textbf{\textit{SIFT1M}}} & \multicolumn{3}{c|}{\textbf{\textit{Deep1M}}}\\
		\hline
		Code\hfill Instr. Set& & Instr. & Dist.& 64 bits & 128 bits & 256 bits & 64 bits & 128 bits & 256 bits \\\hline
		\textbf{\pq{m}{8}~\cite{Jegou2011,FAISS})}&\ref{llpq}&&\dtype{float}&
		\result{22.5}{91.7}{5.42} & \result{44.4}{99.7}{9.20} & \result{62.7}{100.0}{17.2} &
		\result{4.40}{51.4}{5.42} & \result{13.5}{84.6}{9.18} & \result{31.2}{100.}{17.2} \\\hline  
		\textbf{Poylsemous~\cite{Douze2016,FAISS})*}&\ref{llpoly}&\simd{popcnt}&\dtype{bin/f.}&
		\result{22.2}{84.9}{1.22} & \result{44.4}{97.4}{2.25} & \result{60.9}{93.4}{4.15}  & \result{4.80}{40.0}{0.96} & \result{13.3}{76.6}{2.02} & \result{30.6}{93.3}{3.60}  \\
		\sspace Hamming-only ($\tau\!\!=\!\!0$)&& \dtype{popcnt} & \dtype{bin.} &
		\result{0.00}{0.00}{0.87} & \result{0.00}{0.00}{1.30} & \result{0.00}{0.00}{3.01} & \result{0.00}{0.00}{0.85} & \result{0.00}{0.00}{1.41} & \result{0.00}{0.00}{2.97} \\\hline 
		\textbf{Quicker ADC}&&&&&&&&& \\  
		\sspace\ipq{m}{4,\!4}\hfilll SSE && \simd{pshufb} & \dtype{int8} & \result{15.5}{80.6}{0.68} &  \result{30.6}{96.0}{1.25} &  \result{46.5}{99.8}{4.26} &  \result{2.9}{41.8}{0.67} &  \result{14.0}{78.2}{1.32} &  \result{30.0}{97.3}{4.28} \\  
		\sspace\ipq{m}{4,\!4}\hfilll SSE && \simd{pshufb} & \dtype{uint8} &  \result{15.5}{80.9}{0.69} &  \result{31.0}{96.3}{1.21} &  \result{49.9}{99.9}{4.40} &  \result{3.0}{41.5}{0.68} &  \result{14.8}{79.8}{1.31} &  \result{32.4}{97.1}{4.45} \\  
		\sspace\ipq{m}{4,\!4}\hfilll AVX2 && \simd{pshufb} & \dtype{int8}& \result{15.5}{80.6}{0.51} &  \result{30.6}{96.0}{0.83} &  \result{46.5}{99.8}{2.48} &  \result{2.9}{41.8}{0.51} &  \result{14.0}{78.2}{0.92} &  \result{30.0}{97.3}{2.45} \\  
		\sspace\ipq{m}{4,\!4}\hfilll AVX2 && \simd{pshufb} & \dtype{uint8} & \result{15.5}{80.9}{0.54} &  \result{31.0}{96.3}{0.85} &  \result{49.9}{99.9}{2.43} &  \result{3.0}{41.5}{0.54} &  \result{14.8}{79.8}{0.95} &  \result{32.4}{97.1}{2.48} \\  
		\sspace\ipq{m}{4,\!4}\hfilll AVX512 BW &&\simd{pshufb} & \dtype{int8} & \result{15.5}{80.6}{0.49} &  \result{30.6}{96.0}{0.78} &  \result{46.5}{99.8}{2.26} &  \result{2.9}{41.8}{0.49} &  \result{14.0}{78.2}{0.86} &  \result{30.0}{97.3}{2.17} \\  
		\sspace\ipq{m}{4,\!4}\hfilll AVX512 BW &\ref{ll44}&\simd{pshufb} & \dtype{uint8} & \result{15.5}{80.9}{0.53} &  \result{31.0}{96.3}{0.80} &  \result{49.9}{99.9}{2.28} &  \result{3.0}{41.5}{0.52} &  \result{14.8}{79.8}{0.90} &  \result{32.4}{97.1}{2.26} \\  
		\sspace\ipq{m}{4,\!4,\!4,\!4}\hfilll AVX512 && \simd{vpermw} & \dtype{uint16} & \result{15.7}{80.9}{0.84} &  \result{31.6}{96.5}{1.63} &  \result{51.2}{100.}{3.98} &  \result{3.1}{42.1}{0.84} &  \result{14.6}{79.9}{1.73} &  \result{33.6}{97.5}{3.87} \\  
		\sspace{}Quick ADC\cite{Andre2015}\hfilll AVX2&& \simd{pshufb} & \dtype{int8} & \result{15.2}{80.2}{0.51} &  \result{30.1}{95.7}{0.82} &  \result{46.2}{99.7}{2.42} &  \result{3.1}{40.9}{0.5} &  \result{13.7}{77.1}{0.91} &  \result{27.2}{95.3}{2.36} \\  
		\sspace{}Bolt\textsuperscript{16}\cite{Blalock2017}\hfilll AVX2 && \simd{pshufb} & \dtype{uint8} & \result{12.8}{77.0}{0.73} &  \result{29.7}{95.6}{1.34} &  \result{50.7}{100.}{3.45} &  \result{3.1}{41.8}{0.73} &  \result{14.6}{80.3}{1.41} &  \result{33.3}{97.5}{3.42} \\  
		\sspace{}Bolt\textsuperscript{8}\cite{Blalock2017}\hfilll AVX2 && \simd{pshufb} & \dtype{uint8} & \result{12.5}{66.7}{0.46} &  \result{15.0}{32.2}{0.72} &  \result{8.5}{12.2}{2.28} &  \result{0.1}{0.5}{0.43} &  \result{0}{0}{0.82} &  \result{0}{0}{2.30} \\  
		\sspace\ipq{m}{4,\!4} && \simd{  } & \dtype{float}& \result{15.7}{80.9}{25.6} & \result{31.7}{96.5}{54.0} &  \result{51.2}{100.}{104} &  \result{3.0}{42.1}{25.6} &  \result{14.6}{79.8}{54.0} &  \result{33.6}{97.5}{104} \\\hline  
		\sspace\ipq{m}{6,\!6,\!4}\hfilll AVX512 BW && \simd{vpermi2w} & \dtype{int16} & \result{17.4}{85.8}{0.51} &  \result{36.8}{98.5}{0.98} &  \result{56.6}{100.}{2.59} &  \result{4.8}{46.5}{0.48} &  \result{13.6}{81.8}{0.98} &  \result{34.4}{98.5}{2.57} \\  
		\sspace\ipq{m}{6,\!6,\!4}\hfilll AVX512 BW &\ref{ll664}& \simd{vpermi2w} & \dtype{uint16} & \result{17.4}{85.8}{0.51} &  \result{36.8}{98.5}{1.07} &  \result{56.7}{100.}{2.57} &  \result{5.0}{46.3}{0.48} &  \result{13.6}{81.8}{1.07} &  \result{34.3}{98.5}{2.58} \\
		\sspace\ipq{m}{6,\!6,\!4}\hfilll  && \simd{  } & \dtype{float} & \result{17.4}{85.8}{18.4} &  \result{36.8}{98.6}{40.5} &  \result{56.5}{100.}{77.9} &  \result{4.8}{46.3}{18.4} &  \result{13.6}{81.8}{40.4} &  \result{34.4}{98.5}{77.9} \\\hline
		\sspace\ipq{m}{6,\!5,\!5}\hfilll AVX512 BW && \simd{vpermi2w} & \dtype{int16} & \result{17.1}{84.5}{0.50} &  \result{37.5}{98.6}{0.97} &  \result{56.8}{100.}{2.54} &  \result{4.4}{46.8}{0.48} &  \result{14.0}{81.1}{0.97} &  \result{33.7}{98.4}{2.5} \\  
		\sspace\ipq{m}{6,\!5,\!5}\hfilll AVX512 BW &\ref{ll655}& \simd{vpermi2w} & \dtype{uint16} & \result{17.2}{84.5}{0.50} &  \result{37.4}{98.6}{1.04} &  \result{56.8}{100.}{2.55} &  \result{4.4}{46.8}{0.47} &  \result{14.0}{81.1}{1.03} &  \result{33.8}{98.5}{2.47} \\ 
		\sspace\ipq{m}{6,\!5,\!5} && \simd{  } & \dtype{float} &\result{17.1}{84.5}{19.3} & \result{37.4}{98.6}{41.6} &  \result{56.8}{100.}{79.5} &  \result{4.4}{46.8}{19.2} &  \result{14.0}{81.1}{41.6} &  \result{33.8}{98.5}{79.6} \\\hline  
		\sspace\ipq{m}{5,\!5,\!5}\hfilll AVX512 BW && \simd{vpermw} & \dtype{ int16} & \result{14.9}{80.5}{0.50} &  \result{34.4}{97.6}{0.95} &  \result{53.2}{99.9}{2.44} &  \result{2.8}{39.3}{0.47} &  \result{12.0}{77.0}{0.95} &  \result{29.7}{98.5}{2.46} \\
		\sspace\ipq{m}{5,\!5,\!5}\hfilll AVX512 BW && \simd{vpermw} & \dtype{ uint16} & \result{14.9}{80.5}{0.5} &  \result{34.4}{97.6}{0.96} &  \result{53.2}{99.9}{2.48} &  \result{2.9}{39.3}{0.47} &  \result{12.0}{77.0}{0.96} &  \result{29.7}{98.5}{2.49} \\ 
		\sspace\ipq{m}{5,\!5,\!5} && \simd{  } & \dtype{float}& \result{14.9}{80.5}{19.3} &  \result{34.4}{97.7}{41.6} &  \result{53.2}{99.9}{79.5} &  \result{2.8}{39.2}{19.1} &  \result{12.0}{77.0}{41.6} &  \result{29.7}{98.5}{79.6} \\\hline  
		\sspace\ipq{m}{8,\!8}\hfilll AVX512 BW && \simd{vpermi2w} & \dtype{int16} & \result{22.5}{91.8}{1.18} &  \result{44.5}{99.7}{2.3} &  \result{62.8}{100.}{4.89} &  \result{4.3}{51.5}{1.19} &  \result{13.5}{84.6}{2.27} &  \result{31.2}{99.0}{4.9} \\  
		\sspace\ipq{m}{8,\!8}\hfilll AVX512 BW &\ref{ll88}& \simd{vpermi2w} & \dtype{uint16} & \result{22.6}{91.8}{1.19} &  \result{44.5}{99.7}{2.35} &  \result{62.8}{100.}{4.9} &  \result{4.4}{51.4}{1.19} &  \result{13.6}{84.6}{2.26} &  \result{31.2}{99.0}{4.91} \\
		\sspace\ipq{m}{8}\hfilll AVX512 VBMI && \simd{vpermi2b} & \dtype{int8}& \result{22.8}{91.8}{$<$ ** } &  \result{43.7}{99.7}{$<$ ** } &  \result{60.2}{100.}{$<$ ** } &  \result{4.3}{51.6}{$<$ ** } &  \result{13.7}{84.6}{$<$ ** } &  \result{30.6}{99.0}{$<$ ** } \\  
		\sspace\ipq{m}{8}\hfilll AVX512 VBMI && \simd{vpermi2b} & \dtype{uint8} & \result{22.3}{91.6}{$<$ ** } &  \result{42.5}{99.5}{$<$ ** } &  \result{55.6}{100.}{$<$ ** } &  \result{4.5}{51.8}{$<$ ** } &  \result{13.0}{85.1}{$<$ ** } &  \result{28.7}{98.1}{$<$ ** } \\   
		\sspace\ipq{m}{8} && \simd{  } & \dtype{float} & \result{22.5}{91.8}{11.8} &  \result{44.4}{99.7}{22.9} &  \result{62.7}{100.}{47.0} &  \result{4.4}{51.4}{11.8} &  \result{13.5}{84.6}{22.8} &  \result{31.2}{99.0}{47.0}\\\hline  
	\end{tabularx}
	\begin{tablenotes}	
		\scriptsize
		\item[*] For SIFT1M, $\tau=21$ for 64-bit, $\tau=51$ for 128-bit and $\tau=99$ for 256-bit. For Deep1M, $\tau=17$ for 64-bit , $\tau=47$ for 128-bit  and $\tau=99$ for 256-bit.
		\item[**] Timings for \ipq{m}{8} cannot be obtained as current processors do not support AVX512 VBMI; recalls are obtained by simulating saturated arithmetic on quantized distances. As explained in Section~\ref{sec:blend}, we expect AVX512 VBMI  \ipq{m}{8} to be significantly (up to 4 times) faster than AVX512 BW \ipq{m}{8,8}. 
	\end{tablenotes}%
\end{threeparttable}%
}
\begin{tikzpicture}\begin{groupplot}[group style={group size=8 by 1, horizontal sep=0.2cm, vertical sep=0.0cm,}, height=3.0cm,width=3.627cm,ymin=0,ymax=10, 
scatter/classes={%
	pq={mark=diamond,draw=blue!70!black,line width=0.75pt,mark size=2.5pt},
	poly={mark=square,draw=red!70!black,line width=0.75pt,mark size=2.5pt},
	q44={mark=triangle,draw=green!70!black,line width=0.75pt,mark size=2.5pt},
	q66={mark=o,draw=white!50!black,line width=0.75pt,mark size=2.5pt},
	q65={mark=star,draw=white!70!black,line width=0.75pt,mark size=2.5pt},
	q88={mark=x,draw=orange,line width=0.75pt,mark size=2.5pt},
	unused={mark=Mercedes star,draw=black,line width=0.75pt,mark size=2.5pt}
}]
		
\scriptsize
	\nextgroupplot[
			legend columns=2,
			legend to name={CommonLegendC},
	xmin=0.02,
	xmax=0.36,
	ytick={0,3,6,9},
	xtick={0.1,0.2,0.3}]
	
	\addplot[scatter,only marks,%
	scatter src=explicit symbolic]%
	table[meta=label] {
		x y label
		0.225 5.42  pq
		0.222 1.22  poly
		0.155 0.53  q44
		0.174 0.51  q66
		0.172 0.50  q65
		0.226 1.19  q88
	};

	\nextgroupplot[
	legend columns=2,
	legend to name={CommonLegendC},
	xmin=0.77,
	xmax=1.03,
	yticklabels={,,},
	ytick={0.8,0.9,1.0}]
	
	\addplot[scatter,only marks,%
	scatter src=explicit symbolic]%
	table[meta=label] {
		x y label
		0.917 5.42  pq
		0.849 1.22  poly
		0.809 0.53  q44
		0.858 0.51  q66
		0.845 0.50  q65
		0.918 1.19  q88
	};
	
	\nextgroupplot[
	legend columns=2,
	legend to name={CommonLegendC},
	xmin=0.22,
	xmax=0.52,
	yticklabels={,,},
	xtick={0.3,0.4,0.5}]
	
	\addplot[scatter,only marks,%
	scatter src=explicit symbolic]%
	table[meta=label] {
		x y label
		0.444 9.20  pq
		0.444 2.25  poly
		0.310 0.80  q44
		0.368 1.07  q66
		0.374 1.04  q65
		0.445 2.3  q88
	};
	
	\nextgroupplot[
	legend columns=2,
	legend to name={CommonLegendC},
	xmin=0.86,
	xmax=1.03,
	yticklabels={,,},
	xtick={0.9,0.95,1.0}]
	
	\addplot[scatter,only marks,%
	scatter src=explicit symbolic]%
	table[meta=label] {
		x y label
		0.997 9.20  pq
		0.974 2.25  poly
		0.963 0.80  q44
		0.985 1.07  q66
		0.986 1.04  q65
		0.997 2.3  q88
	};

	\nextgroupplot[
	legend columns=2,
	legend to name={CommonLegendC},
	xmin=-0.02,
	xmax=0.14,
	yticklabels={,,},
	xtick={0.0,0.1}]
	
	\addplot[scatter,only marks,%
	scatter src=explicit symbolic]%
	table[meta=label] {
		x y label
		0.044 5.42  pq
		0.048 0.96  poly
		0.030 0.52  q44
		0.050 0.48  q66
		0.044 0.47  q65
		0.044 1.19  q88
	};
	
	\nextgroupplot[
	legend columns=2,
	legend to name={CommonLegendC},
	xmin=0.37,
	xmax=0.53,
	yticklabels={,,},
	xtick={0.4,0.45,0.5}]
	
	\addplot[scatter,only marks,%
	scatter src=explicit symbolic]%
	table[meta=label] {
		x y label
		0.514 5.42  pq
		0.400 0.96  poly
		0.415 0.52  q44
		0.463 0.48  q66
		0.468 0.47  q65
		0.514 1.19  q88
	};
	
	\nextgroupplot[
	legend columns=2,
	legend to name={CommonLegendC},
	xmin=0.08,
	xmax=0.22,
	yticklabels={,,},
	xtick={0.1,0.15,0.20}]
	
	\addplot[scatter,only marks,%
	scatter src=explicit symbolic]%
	table[meta=label] {
		x y label
		0.135 9.18  pq
		0.133 2.02  poly
		0.148 0.90  q44
		0.136 1.07  q66
		0.140 1.03  q65
		0.136 2.26  q88
	};
	
	\nextgroupplot[
	legend columns=2,
	legend to name={CommonLegendC},
	xmin=0.68,
	xmax=0.92,
	yticklabels={,,},
	xtick={0.7,0.8,0.9}]
	
	\addplot[scatter,only marks,%
	scatter src=explicit symbolic]%
	table[meta=label] {
		x y label
		0.846 9.18  pq
		0.766 2.02  poly
		0.798 0.90  q44
		0.818 1.07  q66
		0.811 1.03  q65
		0.846 2.26  q88
	};
	\end{groupplot}
	\scriptsize
	\node[anchor=south, rotate=90] (title-y) at ($(group c1r1.north west)!0.5!(group c1r1.south west)-(0.20cm,0.00cm)$) {Time (ms)};
	\node[anchor=north] (title-sift64) at ($(group c1r1.south)!0.5!(group c2r1.south)-(0.00,0.38cm)$) {SIFT 64-bit};
	\node[anchor=north] (title-sift128) at ($(group c3r1.south)!0.5!(group c4r1.south)-(0.00,0.38cm)$) {SIFT 128-bit};
	\node[anchor=north] (title-deep64) at ($(group c5r1.south)!0.5!(group c6r1.south)-(0.00,0.38cm)$) {Deep 64-bit};
	\node[anchor=north] (title-deep128) at ($(group c7r1.south)!0.5!(group c8r1.south)-(0.00,0.38cm)$) {Deep 128-bit};	
	\foreach \x in {c1r1,c3r1,c5r1,c7r1} {
		\node[anchor=north] (label-\x) at ($(group \x.south west)!0.33!(group \x.south east)-(0.00,0.25cm)$) {R@1};
    }
	\foreach \x in {c2r1,c4r1,c6r1,c8r1} {
		\node[anchor=north] (label-\x) at ($(group \x.south west)!0.68!(group \x.south east)-(0.00,0.25cm)$) {R@100};
	}
	\end{tikzpicture}
	\vspace{-3em}
\end{table*}

The 8-bit shuffles (\verb|pshufb|) in AVX2 or AVX-512 BW operate on two or four independent 128-bit lanes. Hence, they do not allow 32 or 64-element lookup tables. Yet, they can increase the throughput by processing more elements per cycle (see Table~\ref{tbl:simdcap}). We use this property in Quicker ADC ($m\stimes\{4,4\}$) by processing multiple groups (i.e., rows) per shuffle instruction (2  for AVX2, 4  for AVX512) rather than just one per call as in SSE. The number of iterations in the computation is thus reduced from $m/g$ to $m/g/2$ (AVX2) or $m/g/4$ (AVX-512). \trev{This also reduces register pressure as fewer registers are needed to store distance tables.}

\section{Evaluation}
\label{sec:eval}

We implemented Quicker ADC in C++ (4K lines of code) and release it as open-source\footnote{Our implementation integrated into FAISS is released under the Clear BSD license at \url{https://github.com/nlescoua/faiss-quickeradc}}. The implementation contains numerous variants:  $m\stimes\{4,4\}$, $m\stimes\{6,6,4\}$,  $m\stimes\{6,5,5\}$, $m\stimes\{5,5,5\}$, $m\stimes\{8,8\}$, $m\stimes\{8\}$\footnote{In our evaluation, we execute this last implementation in a non-SIMD fallback mode as processors with AVX512 VBMI processors are not available yet. The released code, nevertheless, contains the optimized SIMD implementation. \begin{rev} Our implementation, based on templates, allows adding additional/newer instruction sets easily as it automatically generates code regarding the specific memory layout.\end{rev}}. Note that, to allow further experimentation, the released code is highly generic thanks to templates so that adding a \ipq{m}{6,6,2} operating on signed arithmetics requires a single line of code. Training, exhaustive search and non-exhaustive search (IVF, multi-indexes) rely on the implementation of FAISS.

We carry our experiments on Skylake-based servers, which are m5 AWS instances, built around Intel Xeon Platinum 8175 (2.5 GHz, supporting AVX512) with default settings from Amazon. We use g++ compiler version 7.3 with option \texttt{-O3} and enable SSE, AVX, AVX2 and AVX512. For BLAS, we use the Intel MKL 2018. \begin{rev} We focus our evaluation on Intel's processors as they power almost all servers. Given that optimized implementations are key to relevant evaluations, we also release our source code, to ease evaluation of SIMD schemes on future processors and newer micro-architectures, and for improved quantization schemes.%
\end{rev}%

Our evaluation relies on the publicly available datasets SIFT1M~\cite{Jegou2011} and SIFT1000M~\cite{Jegou2011R} of 128-dimension SIFT vectors, Deep1M~\cite{Babenko2015TQ} of 256-dimension of Deep features, and Deep1B~\cite{Babenko2016} of 96-dimension Deep features. We use the 1-million vectors datasets for the evaluation of exhaustive search, and the 1 billion vectors datasets for the evaluation of non-exhaustive search (i.e., with an index). 

Our metrics are Recall@1 (R@1), which is the fraction of queries for which the true neighbor is the one returned during search, and Recall@100 (R@100), which is the fraction of queries for which the true neighbor is among the top-100 returned during search. R@100 reflects the performance for visual search applications where the user is presented a collection of images rather than a single image (e.g., Google Image). To evaluate the computational efficiency, we report the average time per query. Note that a query time of 0.5ms translates to a throughput of 2,000 queries/second (/core).

\vspace{-0.2em}
\subsection{Exhaustive search}
\label{eval:exhaust}
We first focus on evaluating the performance of Quicker ADC in isolation. We thus consider exhaustive search on the SIFT1M and Deep1M datasets. We do not use an inverted index and we encode the original vectors, not residuals, into short codes. Table~\ref{tab:exhaust} gives results for the baseline product quantization implementation from FAISS~\cite{JDH17}, polysemous codes~\cite{Douze2016}, \trev{Bolt~\cite{Blalock2017},  Quick ADC~\cite{Andre2015}} and Quicker ADC. We also include a specific operating point where polysemous codes degenerate into binary codes ($\tau=0$). 

For Quick ADC~\cite{Andre2015} and Quicker ADC, we scan $t{=}400$ vectors to set the $q_\mathtt{max}$ bound for the quantization of lookup tables (Section 3.4). We evaluate various SIMD-implementations  and compare them to sequential implementations that use floating-point distances to evaluate the loss of recall due to distance quantization.%

\pgfplotscreateplotcyclelist{mycolorlist}{%
	blue!70!black,thick,mark=none\\
	red!70!black,thick,densely dotted,mark=none\\
	green!70!black,thick, densely dashed,mark=none\\
	gray,thick, densely dash dot dot,mark=none\\
	orange,thick, densely dash dot,mark=none\\
}

\pgfplotsset{perf r1/.style={
		xmax=1.0,xtick distance = 0.5,
		ymax=0.32,ytick distance = 0.1,
		restrict x to domain=0:2
}}

\pgfplotsset{perf r100/.style={
		xmax=1.0,xtick distance = 0.5,
		ymax=1.0,ytick distance = 0.5,
		restrict x to domain=0:2
}}

\newcommand{\plotsra}[3]{
	\addplot table[y=R#3,x=time] {plots_recall/#1_#2,PQ8np_NP-r#3.txt.txtb};
	\addplot table[y=R#3,x=time] {plots_recall/#1_#2,PQ8-r#3.txt.txtb};
	\addplot table[y=R#3,x=time] {plots_recall/#1_#2,VPQ_16x4.4_AVX2-r#3.txt.txtb};
	\addplot table[y=R#3,x=time] {plots_recall/#1_#2,VPQ_12x6.6.4_AVX512-r#3.txt.txtb};
	\addplot table[y=R#3,x=time] {plots_recall/#1_#2,VPQ_8x8.8_AVX512-r#3.txt.txtb};
}

\newcommand{\plotsrb}[3]{
	\addplot table[y=R#3,x=time] {plots_recall/#1_#2,PQ16np_NP-r#3.txt.txtb};
	\addplot table[y=R#3,x=time] {plots_recall/#1_#2,PQ16-r#3.txt.txtb};
	\addplot table[y=R#3,x=time] {plots_recall/#1_#2,VPQ_32x4.4_AVX2-r#3.txt.txtb};
	\addplot table[y=R#3,x=time] {plots_recall/#1_#2,VPQ_24x6.6.4_AVX512-r#3.txt.txtb};
	\addplot table[y=R#3,x=time] {plots_recall/#1_#2,VPQ_16x8.8_AVX512-r#3.txt.txtb};
}

\begin{figure*}
	\begin{tikzpicture}\begin{groupplot}[group style={group size=6 by 2, horizontal sep=0.4cm, vertical sep=0.5cm,},
	height=4cm,width=4cm, cycle list name=mycolorlist, ymin=0,xmin=0,
	]
	
	\nextgroupplot[title={\shortstack[c]{IVF (K=65536)\\64-bit codes}}, 
	legend columns=5, 	
	legend to name={CommonLegend},
	perf r1
	]
	
	\plotsra{SIFT1000M}{IVF65536}{1}
	

	\addlegendentry{PQ $m\stimes{}8$}
	\addlegendentry{Polysemous $m\stimes{}8$}
	\addlegendentry{QADC $m\stimes\{4,4\}$}
	\addlegendentry{QADC $m\stimes\{6,6,4\}$}
	\addlegendentry{QADC $m\stimes\{8,8\}$}

	\nextgroupplot[title={\shortstack[c]{IMI (K=$4096^2$)\\64-bit codes}},
	perf r1,
	yticklabels={,,}]
	
	\plotsra{SIFT1000M}{IMI2x12}{1}
	
	\nextgroupplot[title={\shortstack[c]{IVF HNSW (K=$2^{18}$)\\64-bit codes}},
	perf r1,
	yticklabels={,,}]
	
	\plotsra{SIFT1000M}{IVF262144_HNSW32}{1}

	\nextgroupplot[title={\shortstack[c]{IVF (K=65536)\\128-bit codes}},
	perf r1,
	yticklabels={,,}]
	
	\plotsrb{SIFT1000M}{IVF65536}{1}

	\nextgroupplot[title={\shortstack[c]{IMI (K=$4096^2$)\\128-bit codes}},
	perf r1,
	yticklabels={,,}]
	
	\plotsrb{SIFT1000M}{IMI2x12}{1}
	
	\nextgroupplot[title={\shortstack[c]{IVF HNSW (K=$2^{18}$)\\128-bit codes}},
	perf r1,
	yticklabels={,,}]
	
	\plotsrb{SIFT1000M}{IVF262144_HNSW32}{1}
	
	\nextgroupplot[perf r100 ]
	
	\plotsra{SIFT1000M}{IVF65536}{100}

	\nextgroupplot[perf r100, yticklabels={,,}]
	
	\plotsra{SIFT1000M}{IMI2x12}{100}
	
	\nextgroupplot[perf r100, yticklabels={,,}]
	
	\plotsra{SIFT1000M}{IVF262144_HNSW32}{100}
	
	\nextgroupplot[perf r100, yticklabels={,,} ]
	
	\plotsrb{SIFT1000M}{IVF65536}{100}
	
	\nextgroupplot[perf r100, yticklabels={,,}]
	
	\plotsrb{SIFT1000M}{IMI2x12}{100}
	
	\nextgroupplot[perf r100, yticklabels={,,}]
	
	\plotsrb{SIFT1000M}{IVF262144_HNSW32}{100}
	
	\end{groupplot}
	
	\node[anchor=north] (title-x) at ($(group c1r2.south east)!0.5!(group c6r2.south west)-(0,0.4cm)$) {Query time(ms)};
	\node[anchor=south, rotate=90] (title-ya) at ($(group c1r1.west)-(0.50,0cm)$) {Recall@1};
	\node[anchor=south, rotate=90] (title-yb) at ($(group c1r2.west)-(0.50,0cm)$) {Recall@100};
	
	\node[below=-0.1cm of title-x] (legend) {\ref{CommonLegend}};%
	\end{tikzpicture}%
	\caption{Non-exhaustive search with different index types and different PQ or Irregular PQ codes. SIFT1000M}
	\label{fig:exhaust_sift1000m}
\end{figure*}
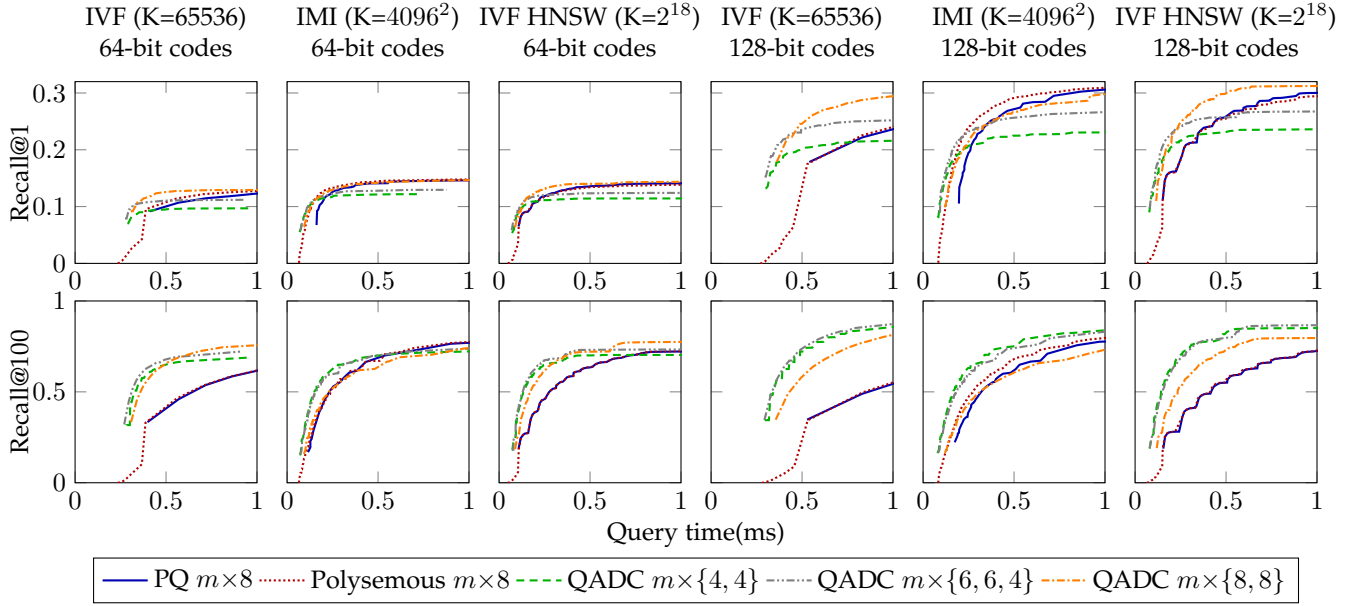%
\begin{rev}%
	\vspace{-0.2em}
\subsubsection{Impact of distance quantization}
Distance quantization has no impact on recall with 16-bit integers and negligible impact with 8-bit integers (e.g., 0.155 instead of 0.157 for \ipq{m}{4,\!4} for R@1 SIFT1M - 64 bit). Hence, scanning $t{=}400$ vectors is enough for estimating the distance-quantization bounds. The unsigned variants have a slight recall advantage over the signed ones.  As query times for both are similar, we'll keep the unsigned variants. 

Quicker ADC \ipq{m}{4,4} with 8-bit distance is faster than \ipq{m}{4,4,4,4} with 16-bit distances for similar recall. We thus prefer using 8-bit distances whenever an appropriate shuffle is available. Indeed, shuffling 16-bit values has a higher latency (see Table~\ref{tab:count}), converting 8-bit to 16-bit has a cost and the number of additions per cycle is divided by 2. 

Regarding prior work, Quicker ADC \ipq{m}{4,4} (AVX2) improves over Quick ADC~\cite{Andre2015}, which also uses AVX2, with a better recall for the same query time thanks to the tighter $q_\mathtt{min}$ bound. Quicker ADC outperforms Bolt\textsuperscript{16} by providing better recall (thanks to the different distance quantization scheme) and by being faster (thanks to accumulating on 8-bit instead of upcasting partial distances to 16-bits). Note that Bolt\textsuperscript{8} with an 8-bit accumulator has poor recalls as the quantization bounds are estimated on partial distances rather than summed distance, thus a lot of relevant distances saturate during accumulation. 

As a side note, our sequential implementation tends to be slower than original the PQ ADC because it is not as specialized and it systematically uses shifting and masking for accessing subcodes. For \ipq{m}{4,4}, using SSE is slower than using AVX2 or AVX512 which give similar timings. 

\subsubsection{Comparing QuickerADC to PQ and Polysemous}
Quicker ADC \ipq{m}{6,6,4} is as fast as \ipq{m}{4,4} yet improves recall for both 64-bit and 128-bit codes. Its recall R@1 (0.174 and 0.368 on SIFT1M 64-bit/128-bit) is lower than PQ and polysemous codes (0.225 and 0.444) but the recall R@100 of Quicker ADC is equivalent if not better than the one of polysemous codes. Yet, this slight recall decrease allows Quicker ADC  \ipq{m}{6,6,4} to be 10 times faster than PQ ADC and 2-3 times faster than polysemous codes.
Interestingly, \ipq{m}{5,5,5} has a lower recall than all other including \ipq{m}{4,4} because it uses only 60 bits instead of all 64 bits. Hence, irregular product quantizers are preferable over simpler constructions not using all bits.

Quicker ADC $m\stimes{\{8,8\}}$ codes are 5$\stimes$ faster than regular PQ codes, and slightly faster than polysemous codes while achieving similar recall R@1 and improved recall R@100. Note that the timings reported here are using AVX512 BW and are likely to be up to 4$\stimes$ better with the availability of AVX512 VBMI without significant degradation of accuracy. Thus, \ipq{m}{8} codes with Quicker ADC could prove extremely interesting due to their recall being very close to the original PQ with significantly improved performance. 

Quicker ADC is faster than a binary code (polysemous with $\tau=0$) as the shuffle instructions used in Quicker ADC are faster than \texttt{popcount}. This opens perspectives for using PQ with Symmetric Distance Computation with an SIMD implementation similar to that of Quicker ADC in order to build a very fast code that could be used in pruning applications (e.g., in place of the hamming code of polysemous codes). Bounding a \pq{m}{8} with a \pq{m}{4} code is one of the two mechanisms used in~\cite{Andre2015}; it could be used in isolation as explained in Derived Quantizers~\cite{Thesis}, in a way similar to the hamming codes of polysemous codes~\cite{Douze2016}.\end{rev}%

\pgfplotsset{perf r1/.style={%
		xmax=1.0,xtick distance = 0.5,%
		ymax=0.4,ytick distance = 0.2,%
		restrict x to domain=0:40%
}}%
\begin{figure*}%
	\begin{tikzpicture}\begin{groupplot}[group style={group size=6 by 2, horizontal sep=0.4cm, vertical sep=0.5cm,},
	height=4cm,width=4cm, cycle list name=mycolorlist, ymin=0,xmin=0,
	]
	
	\nextgroupplot[title={\shortstack[c]{IVF (K=65536)\\64-bit codes}}, 
	legend columns=5, 	
	legend to name={CommonLegend2},
	perf r1
	]
	
	\plotsra{Deep1B}{IVF65536}{1}
	

	\addlegendentry{PQ $m\stimes{}8$}
	\addlegendentry{Polysemous $m\stimes{}8$}
	\addlegendentry{QADC $m\stimes\{4,4\}$}
	\addlegendentry{QADC $m\stimes\{6,6,4\}$}
	\addlegendentry{QADC $m\stimes\{8,8\}$}

	\nextgroupplot[title={\shortstack[c]{IMI (K=$4096^2$)\\64-bit codes}},
	perf r1,
	yticklabels={,,}]
	
	\plotsra{Deep1B}{IMI2x12}{1}
	
	\nextgroupplot[title={\shortstack[c]{IVF HNSW (K=$2^{18}$)\\64-bit codes}},
	perf r1,
	yticklabels={,,}]
	
	\plotsra{Deep1B}{IVF262144_HNSW32}{1}

	\nextgroupplot[title={\shortstack[c]{IVF (K=65536)\\128-bit codes}},
	perf r1,
	yticklabels={,,}]
	
	\plotsrb{Deep1B}{IVF65536}{1}

	\nextgroupplot[title={\shortstack[c]{IMI (K=$4096^2$)\\128-bit codes}},
	perf r1,
	yticklabels={,,}]
	
	\plotsrb{Deep1B}{IMI2x12}{1}
	
	\nextgroupplot[title={\shortstack[c]{IVF HNSW (K=$2^{18}$)\\128-bit codes}},
	perf r1,
	yticklabels={,,}]
	
	\plotsrb{Deep1B}{IVF262144_HNSW32}{1}
	
	\nextgroupplot[perf r100 ]
	
	\plotsra{Deep1B}{IVF65536}{100}

	\nextgroupplot[perf r100, yticklabels={,,}]
	
	\plotsra{Deep1B}{IMI2x12}{100}
	
	\nextgroupplot[perf r100, yticklabels={,,}]
	
	\plotsra{Deep1B}{IVF262144_HNSW32}{100}
	
	\nextgroupplot[perf r100, yticklabels={,,} ]
	
	\plotsrb{Deep1B}{IVF65536}{100}
	
	\nextgroupplot[perf r100, yticklabels={,,}]
	
	\plotsrb{Deep1B}{IMI2x12}{100}
	
	\nextgroupplot[perf r100, yticklabels={,,}]
	
	\plotsrb{Deep1B}{IVF262144_HNSW32}{100}

	\end{groupplot}
	
	\node[anchor=north] (title2-x) at ($(group c1r2.south east)!0.5!(group c6r2.south west)-(0,0.5cm)$) {Query time(ms)};
	\node[anchor=south, rotate=90] (title2-ya) at ($(group c1r1.west)-(0.50,0cm)$) {Recall@1};
	\node[anchor=south, rotate=90] (title2-yb) at ($(group c1r2.west)-(0.50,0cm)$) {Recall@100};
	
	\node[below= 0.0cm of title2-x] (legend2) {\ref{CommonLegend2}};
	
	\end{tikzpicture}%
	\caption{Non-exhaustive search with different index types and different PQ or Irregular PQ codes. Deep1B}%
	\label{fig:exhaust_deep1b}%
\end{figure*}
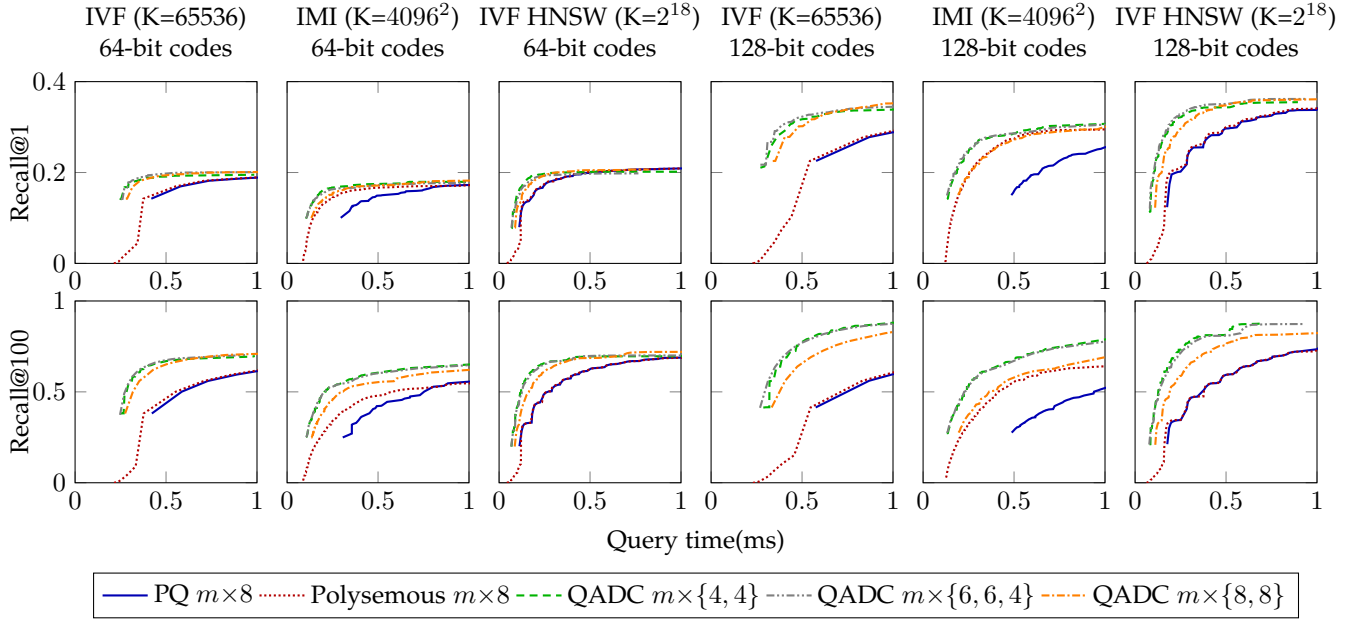
\subsection{Non-exhaustive search}
\label{sec:evalexhaust}
Each index and product quantizer combination can operate in numerous configurations by varying the hyper-parameters (i.e., How many cells to explore? How many distances to evaluate?  What hamming threshold to use? How to estimate distance quantizers?). Hence, the performance of each combination is not a single point but a curve of the optimal tradeoffs between query time (in ms) and recall (R@1 or R@100). Thus, to compare the various combinations of indexes and product quantizers, we plot the best recall achieved for a given query time budget. We are particularly interested by small query times (less than 0.5ms). \trev{Among the parameters of importance is the parameter $a$ (multiple assignement/probe) which allows faster codes to scan more vectors and more inverted lists for a given time budget.}

We consider 3-types of indexes combined with either 64-bit or 128-bit codes. The first index considered is a relatively coarse index (IVF $K=65536$)~\cite{Jegou2011}, the second is a very fine-grained index (inverted-multi-index, IMI $K=4096^2$)~\cite{Babenko2012}, and the last one is a fine-grained index leveraging a neighborhood graph (IVF HNSW $K=2^{18}$)~\cite{Malkov2016}. The two latters are considered state-of-the-art and are widely used; they achieve similar performance and the more recent IVF HNSW produces fewer inverted lists, which are also more balanced. We report results for the SIFT1000M dataset~\cite{Jegou2011R} on Figure~\ref{fig:exhaust_sift1000m} and for the Deep1B dataset~\cite{Babenko2016}) on Figure~\ref{fig:exhaust_deep1b}.  Note that gaps in the curves (quite visible for 128-bit PQ and polysemous codes with IVF HNSW) are related to the fact that parametrizations are discrete (i.e., exploring 1,2,3,... inverted lists).

For all indexes, both codes (64-bit and 128-bit) and both datasets (SIFT1000M and Deep1B), the top performer is one of Quicker ADC's implementations, with a single exception (SIFT1000M, IMI, 128-bit codes for the metric R@1, with a budget for query time $>$ 0.2 ms). Indeed, as observed in Section~\ref{sec:evalexhaust}, polysemous codes tend to perform better on R@1; they also avoid distance table computation which are numerous in IMI. The domination of Quicker ADC is particularly salient for low query time budget, where the recall gain of Quicker ADC over polysemous codes can be 50\% (SIFT100M, IMI, 128-bit, 0.2 ms) or 100\% (SIFT 1000M, IVF HSNW, 128-bit, 0.2ms). Quicker ADC is also particularly efficient for metric R@100, or for Deep 1B vectors. For example, on Deep1B with an IMI 2x12 with a query time budget of 0.25ms, Quicker ADC \ipq{32}{4,4} allows a R@100 of 0.55 while polysemous allow a R@100 of 0.33 or would require a query time of 0.5ms to achieve the same R@100, and PQ \rpq{16}{8} requires more than 1ms to achieve a recall R@100 of 0.55. Quicker ADC becomes particularly interesting when combined with IVF HNSW: for example, \ipq{32}{4,4} achieves a R@100 of 0.55 in less than 0.16ms; faster than alternatives on IMI. Indeed, IMI is a worst case for Quicker ADC as these indexes tend to have numerous short lists, and Quicker ADC requires (i) pre-computing distance tables, and (ii) processing batches of vectors. The benefits are higher for indexes that have longer inverted lists (IVF, IVF HNSW). \trev{The reduced accuracy of QADC \ipq{m}{4,\!4} or \ipq{m}{6,\!6,4},\! observed on exhaustive searched, is here offset by the ability to scan more inverted lists (larger hyper-parameter $a$) for a given time budget. This explains why QADC outperforms PQ and polysemous codes in term of accuracy.}%

Note that for index types other than IMI, polysemous codes allow additional operating points over PQ (lower query time by jeopardizing recall) but do not improve performance for the larger query times.  It means that while the reduced cost per code of polysemous allows computing distances for more codes/more inverted lists, an equally effective alternative is to not use polysemous but scan fewer codes/less inverted lists.  The fact that they are more benefical in the context of IMI can be explained by the fact that they allow avoiding distance table computations, that are more numerous for such fine-grained indexes. Interestingly, their benefits are limited on IVF HNSW.


\subsection{Summary of experimental study}

Our experiments show that Quicker ADC offers interesting operating points for exhaustive search and non-exhaustive search, even with fine-grained indexes. For exhaustive search, new variants relying on AVX512 are the fastest, with \ipq{m}{8,8} offering good recall for both R@1 and R@100 with significantly reduced query times. For non-exhaustive search, Quicker ADC \ipq{m}{4,4} or \ipq{m}{6,6,4} outperforms other solutions, including polysemous when fast queries are desired. As the time budget for queries increases, or when R@1 is the metric of interest, Quicker ADC\ipq{m}{8,8} is among the top performers. Hence, Quicker ADC \ipq{m}{8} whose ADC procedure could be up to 4$\stimes$ faster (see Table~\ref{tab:count}) is likely to be well positioned when processors supporting the required instructions will become available.

\section{Related Work}
\emph{SIMD evaluation of distances.} PQ Fast Scan~\cite{Andre2015} pioneered the use of SIMD for ADC distance evaluation. It inspired later work~\cite{Andre2017,Blalock2017, Wu2017} that are more suitable for indexed databases. 
\begin{rev} Quicker ADC is a generalization of Quick ADC~\cite{Andre2017} supporting additional shuffles and an improved distance quantization scheme. Bolt~\cite{Blalock2017} is cotemporary of Quick ADC and both implement lookups using \texttt{pshufb}. Bolt however differs in the approach of distance quantization. In Bolt, the distance quantization scheme is decided once for all at training time and aims at minimizing distance distortion for all distances. In Quick ADC and Quicker ADC, the distance quantization scheme is decided dynamically at query-time based on the first vectors scanned and aims at minimizing distortion for only the $k$-closest vectors, quantizing distances for all other vectors to $\infty$. This also implies different choices for accumulation: Bolt quantizes distances to 8-bit and accumulates on 16-bit whereas Quick ADC and Quicker ADC quantize distances to 8-bit and accumulate them on 8-bit with saturated arithmetic to ensure semantic coherence between $\infty$ encoded onto 8-bit in tables and $\infty$ encoded onto 8-bit during accumulation --- should we accumulate on 16-bit, 8-bit sub-distances quantized to $\infty$ wouldn't be interpreted as $\infty$ during accumulation. Quick ADC and Quicker ADC prove more effective for nearest neighbor search, while Bolt allows more general use in other applications (matrix multiplication\ldots). Wu et al. \cite{Wu2017} present an improved method for learning product quantizers and depart from the usual \rpq{m}{8} to also evaluate \rpq{m}{4} variants, which can be implemented in SIMD. They report execution speed 5$\stimes$ faster, and equaling \texttt{popcnt}-based hamming distances speed, similar to our observations. These propositions~\cite{Andre2015,Blalock2017,Wu2017} have shown to be particularly efficient yet their evaluations remains limited to no or coarse indexes and thus lack results for fine grained indexes (e.g., Inverted Multi Index~\cite{Babenko2015}). 
\end{rev}

By implementing Quicker ADC into FAISS, we have been able to evaluate Quicker ADC with both IMI and IVF HNSW. Also, as we release the implementation, it will be possible to evaluate Quicker ADC for future index designs. In addition, these works were limited to \rpq{m}{4} codes, while we have proposed numerous other variants for a better usage of AVX512-compatible processors.

\emph{Indexes.} Inverted Multi-Indexes~\cite{Babenko2015} or graph-based inverted indexes \cite{Malkov2016,Baranchuk_2018_ECCV} provide a finer partition than vanilla inverted indexes. The design of indexes is rather independent from the design of the product quantizer, and thus Quicker ADC can be combined with any inverted index. We evaluate Quicker ADC with several types of indexes and shows that it works fine with IMI which are at the extreme of spectrum in term of cell sizes and works equally well, contrary to polysemous codes with more recent propositions such as IVF HNSW. We expect Quicker ADC to be relatively independent of index choices and to remain efficient with newer indexes.

\begin{rev}
Alternatively, PQTable\cite{Matsui2018} seeks to avoid the distance computation for numerous vectors by using a hash-table-like structure, using a PQ code as the key. While it's as fast as less optimized implementations of PQ with Inverted Multi Indexes~\cite{Babenko2015}, the evaluation reports 7 ms to achieve 0.06 R@1 and 0.57 R@100 which are much lower than speeds achieved by FAISS using polysemous~\cite{Douze2016} and Quicker ADC ($< 0.3$ ms in Figure~\ref{fig:exhaust_sift1000m}). Yet, both polysemous using \texttt{popcount} and Quicker ADC are optimized according to hardware capabilities, and PQTable could also benefit from additional CPU-aware optimization, using batching and prefetching similarly to classical hash-tables~\cite{LeScouarnec2017}. 
\end{rev}

\emph{Optimized and Compositional Quantization Models.} Cartesian k-means (CKM) \cite{Norouzi2013} and Optimized Product Quantizers (OPQ) \cite{Ge2014}  optimize the sub-space decomposition by performing an arbitrary rotation and permutation of vector components. This allows for improved accuracy with a moderate cost (i.e., a matrix multiplication to perform the rotation). These are compatible with Quicker ADC as the ADC procedure remains unchanged. The source code we release includes FAISS's implementation of OPQ. We focused our evaluation on the combination of Quicker ADC with regular PQ with various indexes. In addition, compositional vector quantization models inspired by PQ have been proposed. These models offer a lower quantization error than PQ or OPQ. Among these models are Additive Quantization (AQ) \cite{Babenko2014AQ}, Tree Quantization (TQ) \cite{Babenko2015TQ} and Composite Quantization (CQ) \cite{Zhang2014}. These models also use cache-resident lookup tables to compute distances, therefore Quicker ADC may be combined with them. However, this may require additional work as some of these models use more lookup tables than the ADC procedure of PQ or OPQ.

\emph{Deep-Learning-based quantizers.} Subic~\cite{Jain2017}, DPQ~\cite{Klein2017} and ~\cite{Yu_2018_ECCV} use deep neural networks to compute a compact vector representating images. Similarly to product quantization, the compact vector has a product structure which is exploited to compute distances by summing the contribution of sub-vectors. The distances to each sub-vectors are thus stored in lookup tables and an ADC-like procedure is used. Hence, Quicker ADC naturally extends to these quantizers, and can bring similar benefits. Quicker ADC could be particularly interesting if these quantizers can be adapted to accomodate well small (4-bits) or irregular quantizers. 

\emph{Encodings based on neighborhood graphs.} Some propositions~\cite{Babenko2017,Douze2018} leverage the nearest neighbor graph to have lower encoding error. This improves recall but tend to operate with a higher memory budget~\cite{Douze2018} and thus does not compare directly. Our work target operating points strictly identical to IVF or inverted multi-indexes combined with Product Quantization~\cite{Jegou2011,Babenko2015} and Polysemous Codes~\cite{Douze2016}. Yet, as these propositions ~\cite{Babenko2017,Douze2018} rely on lookup tables for distance computation, they could leverage some principles from Quicker ADC to speed up distance computation.

\section{Conclusion}

In this paper, we presented Quicker ADC, a novel distance computation method for product-quantization-based ANN search. Quicker ADC improves over previous proposition~\cite{Andre2017} by (i) supporting additional quantizers (e.g., \ipq{m}{6,6,4}, \ipq{m}{8,8}, \ipq{m}{8}) and (ii) having an improved implementation integrated into FAISS and compatible with various indexes (IMI, IVF HNSW). Through an extensive evaluation, we have shown that Quicker ADC outperforms schemes based on PQ or polysemous codes for both exhaustive and non-exhaustive (i.e., index-based) search, and that they combine well with the latest indexes such as HNSW-based IVF~\cite{Malkov2016}. We release the implementation as open-source to allow a wider adoption and evaluation of this approach.

Techniques presented in this paper focus on the efficient evaluation of distances in the compressed domain. This problem is present in all quantization-based approaches, which rely on lookup tables to speed-up computation. Thus, the principles behind our work (i.e., replacing memory accesses by shuffles and quantizing distance tables) can be the basis for an improved implementation of other approaches~\cite{Jain2017,Babenko2017,Douze2018}. This is of particular interest if those new approaches behave better with coarser (i.e., 4-bit) or irregular quantizers (i.e., combined 6-bit and 5-bit quantizers). Also, Quicker ADC brings product quantizer codes on par with binary codes regarding distance evaluation speed. Thus, Quicker ADC could inspire new designs where filtering with a binary code (e.g., polysemous codes) is replaced by filtering with a lower precision product quantizer (e.g., \ipq{m}{4,4} with symmetric distance computation used to filter vectors before ADC computation on \pq{m}{8} as in~\cite{Andre2015,Thesis}). 

Finally, we would like to stress that upcoming processors will have improved SIMD capabilities allowing for a bright future for Quicker ADC. The Cannonlake processors expected in 2019 will have support for 7-bit shuffles thus quadrupling shuffle throughput as \ipq{m}{8} codes replace \ipq{m}{8,8}, and Sunny Cove processors expected in 2020 will, in addition, double shuffle throughput by having two shuffle units per core instead of one. Hence, the performance of Quicker ADC will significantly improve in a near future just from hardware upgrade, without algorithm adaptation.

\bibliographystyle{plain}
\bibliography{references}

\section*{Biographies}
\vspace{-2em}
\begin{IEEEbiography}[{\includegraphics[width=1in,height=1.00in,clip,keepaspectratio]{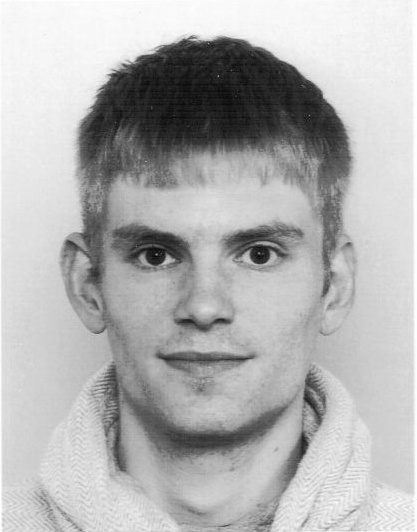}}]{Fabien André} is a software engineer. He worked at Technicolor from 2013 to 2018 on multimedia databases and high-performance networking systems. He earned an engineering degree from INSA de Rennes, France in 2013, and a PhD from INSA de Rennes, France in 2016.
\end{IEEEbiography}
\vspace{-2.5em}

\begin{IEEEbiography}[{\includegraphics[width=1in,height=1.25in,clip,keepaspectratio]{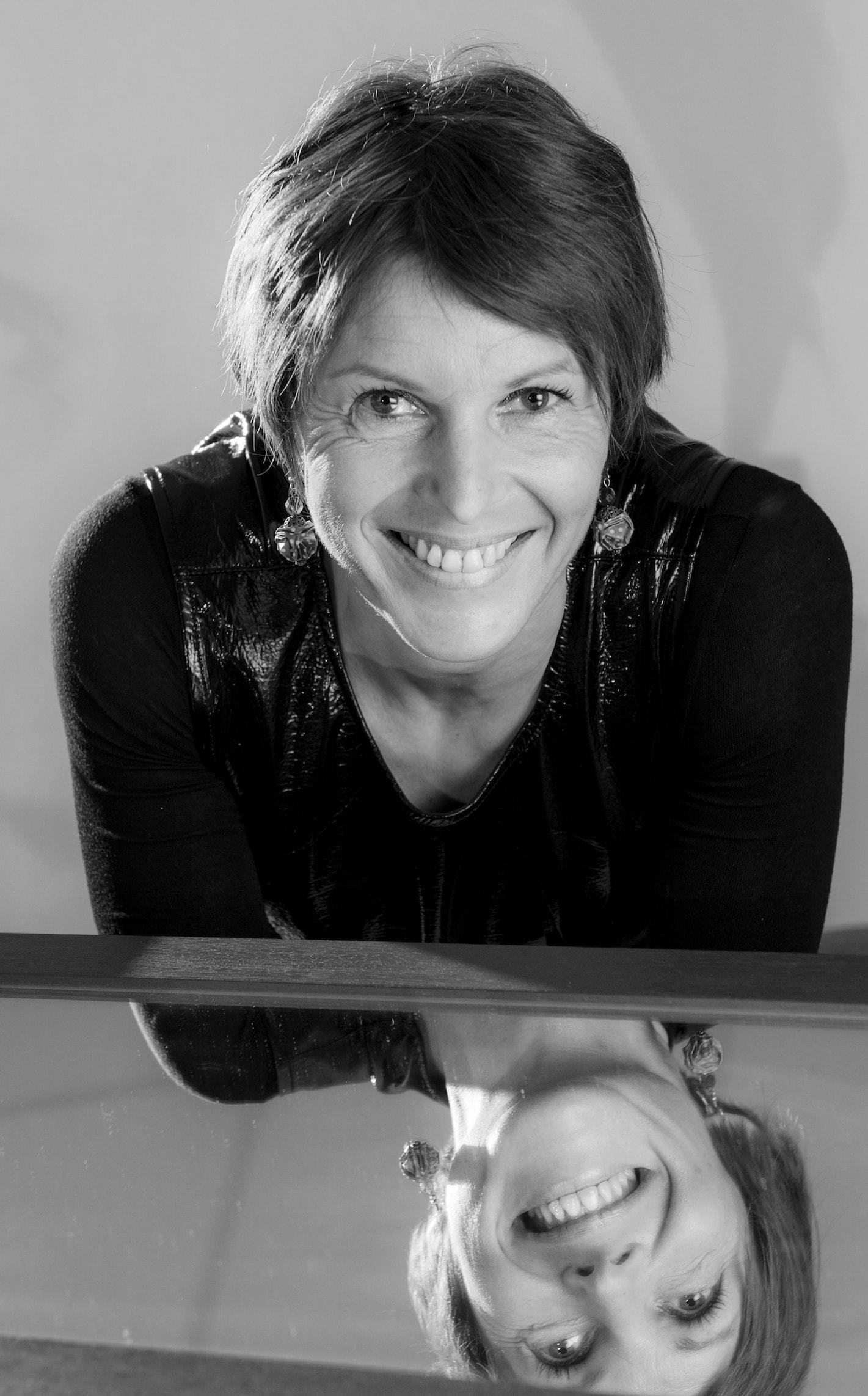}}]{Anne-Marie Kermarrec} is Full Professor at EPFL, Swizerland. She founded the Mediego startup in April 2015. Mediego provides online predictive marketing services that directly leverage her recent research.  Before that, after her PhD thesis at University of Rennes in 1996, she has been with Vrije Universiteit, NL and Microsoft Research Cambridge, UK. She was Research Director at Inria in Rennes from 2004 to 2019. She is an ACM fellow since 2016.
\end{IEEEbiography}
\vspace{-2em}
\begin{IEEEbiography}[{\includegraphics[width=1in,height=1.25in,clip,keepaspectratio]{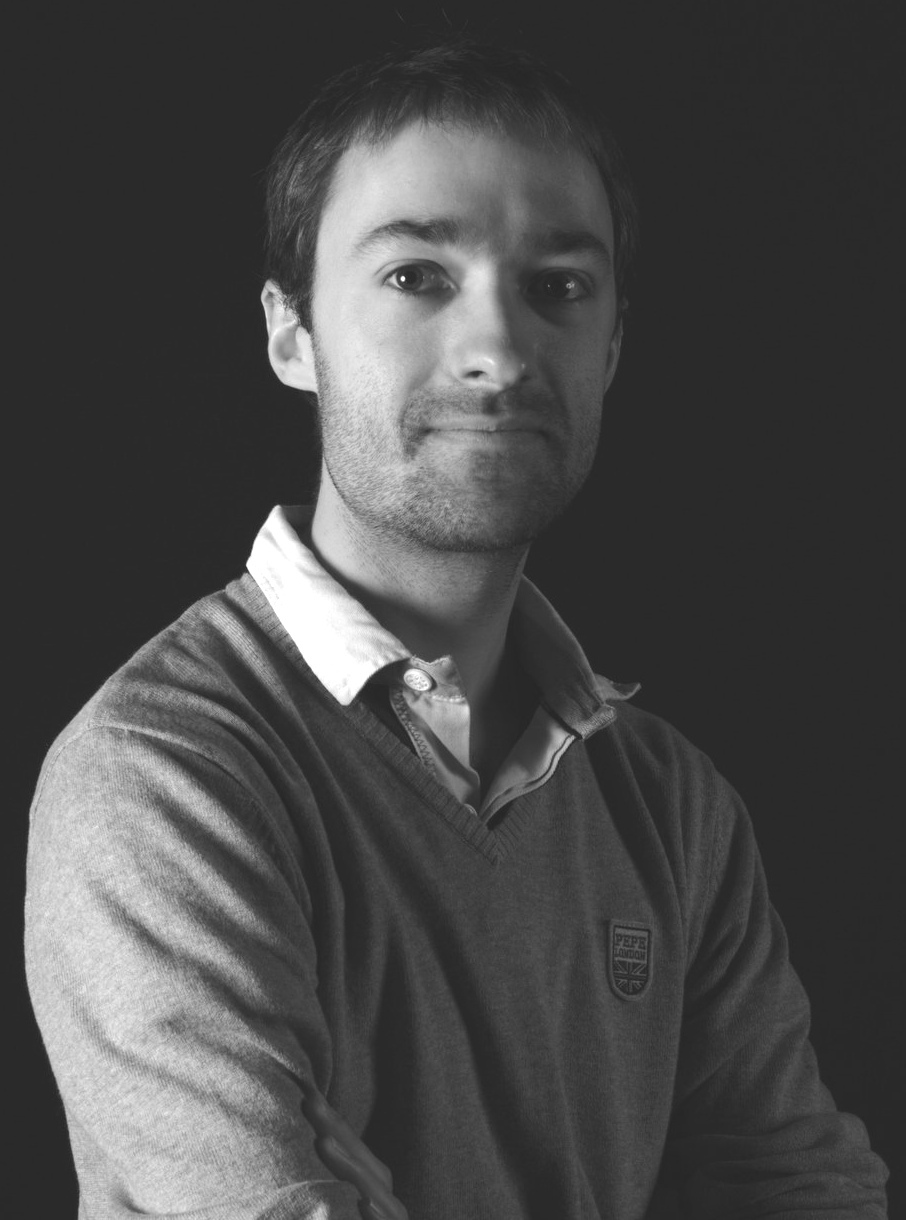}}]{Nicolas Le Scouarnec} is a research engineer at Broadpeak. Before that he was a senior scientist in distributed systems and high performance computing at Technicolor. His current research focuses on reliable and high performance networked systems, as well efficient implementation of machine learning algorithms. He earned a M.Sc. degree from INSA de Rennes in 2007 and a Ph.D. in computer science from INSA de Rennes, France in 2010.
\end{IEEEbiography}
\end{document}